\documentclass[letterpaper, 10 pt, conference]{ieeeconf}  

\IEEEoverridecommandlockouts                              

\let\labelindent\relax
\usepackage{graphicx}
\usepackage{amsmath}
\usepackage{amssymb}
\usepackage{nccmath}
\usepackage{multirow}
\usepackage{algorithmic}
\usepackage[ruled]{algorithm2e}
\usepackage{acronym}
\usepackage{xspace}
\usepackage{enumitem}
\usepackage{hyperref}
\usepackage{balance}
\usepackage{setspace}
\usepackage[font=small,skip=0pt]{subcaption}
\usepackage[font=small,skip=0pt]{caption}
\usepackage[misc]{ifsym}
\usepackage[dvipsnames]{xcolor}
\usepackage[capitalise]{cleveref}
\usepackage{overpic}

\DeclareMathOperator*{\argmax}{arg\,max}

\makeatletter
\DeclareRobustCommand\onedot{\futurelet\@let@token\@onedot}
\def\@onedot{\ifx\@let@token.\else.\null\fi\xspace}
\def\eg{\emph{e.g}\onedot} 
\def\ie{\emph{i.e}\onedot} 
 
\def\etc{\emph{etc}\onedot} 
 
\def\etal{\emph{et al}\onedot}
\makeatother

\makeatletter
\def\BState{\State\hskip-\ALG@thistlm}
\makeatother

\makeatletter
\renewcommand{\paragraph}{%
  \@startsection{paragraph}{4}%
  {\z@}{0ex \@plus 0ex \@minus 0ex}{-1em}%
  {\hskip\parindent\normalfont\normalsize\bfseries}%
}
\makeatother

\crefname{algocf}{alg.}{algs.}
\Crefname{algocf}{Algorithm}{Algorithms}
\newcommand\energy{\mathcal{E}}

\acrodef{tom}[ToM]{Theory of Mind}
\acrodef{mot}[MOT]{Multiple Object Tracking}
\acrodef{mvt}[MVT]{Multiple View Tracking}
\acrodef{pg}[\textit{pg}]{parse graph}
\acrodef{map}[MAP]{maximizing a posterior}

\title{\LARGE \bf 
Joint Inference of States, Robot Knowledge, and Human (False-)Beliefs
}

\author{Tao Yuan$^{1}$\quad{}Hangxin Liu$^{1}$\quad{}Lifeng Fan$^{1}$\quad{}Zilong Zheng$^{1}$\quad{}Tao Gao$^{1,3}$\quad{}Yixin Zhu$^{1,2}$\quad{}Song-Chun Zhu$^{1,2}$\quad{}
\thanks{$^{1}$ UCLA Center for Vision, Cognition, Learning, and Autonomy (VCLA) at Statistics Department. Emails:
{\tt\small \{taoyuan, hx.liu, lfan, z.zheng, yixin.zhu\}@ucla.edu}, \tt\small{\{tao.gao, sczhu\}@stat.ucla.edu}.}%
\thanks{$^{2}$ International Center for AI and Robot Autonomy (CARA)}%
\thanks{$^{3}$ UCLA Department of Communication}%
\thanks{The work reported herein was supported by ONR MURI N00014-16-1-2007 (SCZ), DARPA XAI N66001-17-2-4029 (SCZ), ONR N00014-19-1-2153 (SCZ), and DARPA-PA-19-03-01 (TG).}%
}

\begin{document}

\maketitle
\thispagestyle{empty}
\pagestyle{empty}

\begin{abstract}
Aiming to understand how human (false-)belief---a core socio-cognitive ability---would affect human interactions with robots, this paper proposes to adopt a graphical model to unify the representation of object states, robot knowledge, and human (false-)beliefs. Specifically, a \acf{pg} is learned from a single-view spatiotemporal parsing by aggregating various object states along the time; such a learned representation is accumulated as the robot's knowledge. An inference algorithm is derived to fuse individual \ac{pg} from all robots across multi-views into a joint \ac{pg}, which affords more effective reasoning and inference capability to overcome the errors originated from a single view. In the experiments, through the joint inference over \ac{pg}s, the system correctly recognizes human (false-)belief in various settings and achieves better cross-view accuracy on a challenging small object tracking dataset.

\end{abstract}

\setstretch{0.93}

\section{Introduction}

\begin{figure}[t!]
    \centering
    \includegraphics[width=\linewidth,trim={0.5cm 0.5cm 0.5cm 1cm},clip]{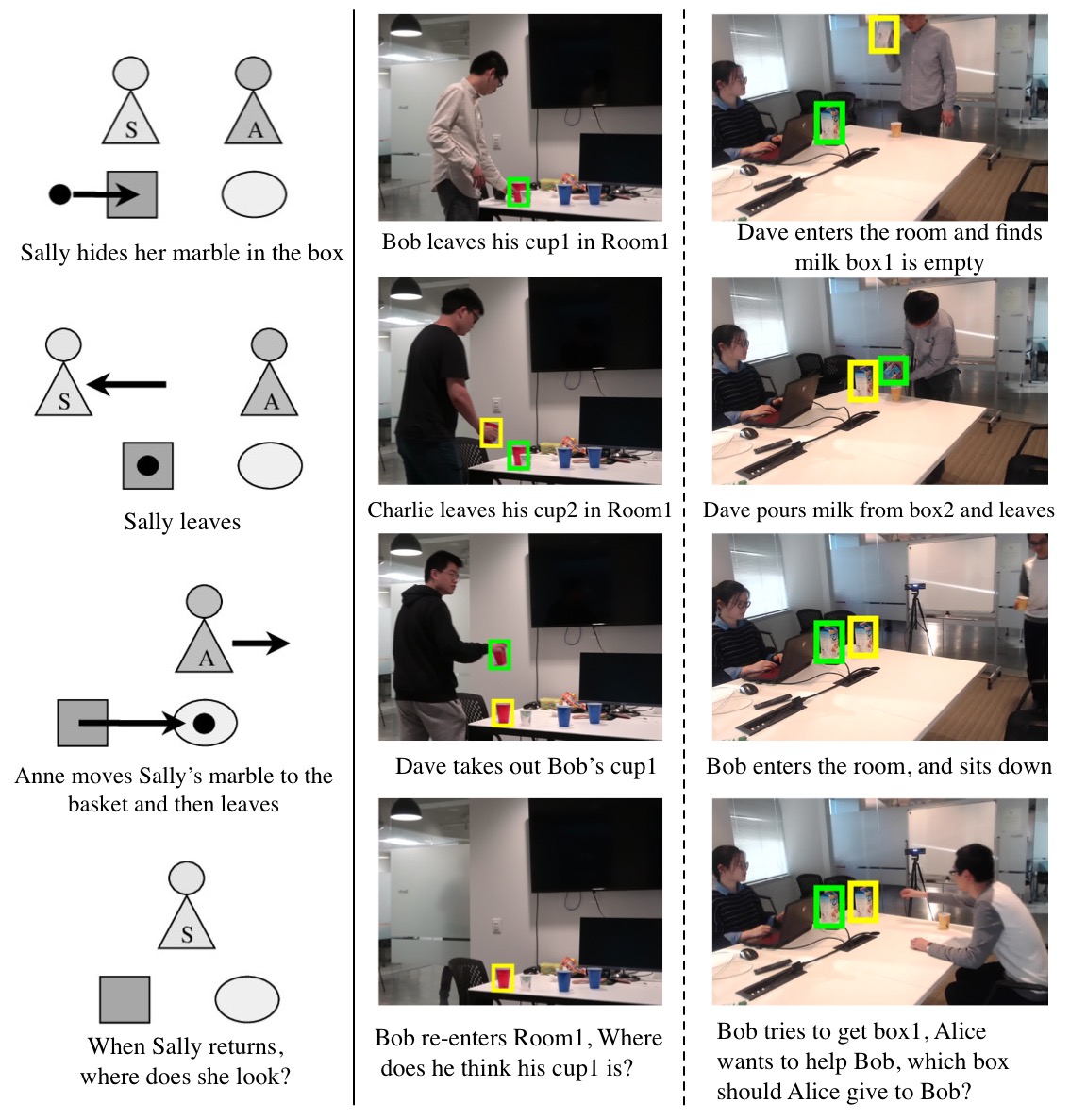}
    \caption{Left: Illustration of the classic Sally-Anne test~\cite{baron1985does}. Middle and Right: Two different types of false-belief scenarios in our dataset: belief test and helping test.}
    \label{fig:sally}
\end{figure}

The seminal Sally-Anne~\cite{baron1985does} study has spawned a vast research literature in developmental psychology regarding \ac{tom}; in particular, human's socio-cognition in understanding \emph{false-belief}---the ability to understand other's belief about the world may contrast with the true reality. A cartoon version of the Sally-Anne test is shown in the left of \cref{fig:sally}: Sally puts her marble in the box and left. While Sally is out, Anne moves the marble from the box to a basket. The test would ask a human participant where Sally would look for her marble when she is back. In this experiment, the marble would still be inside the box according to Sally's false-belief, even though the marble is actually inside the basket. To answer this question correctly, an agent should understand and disentangle the object state (observation from the current frame), the (accumulated) knowledge, the belief of other agents, the ground-truth/reality of the world, and importantly, the concept of false-belief.

The prior study suggests that at the age of 4 years old, children begin to develop the capability to understand false-belief~\cite{gopnik1988children}. Such abilities to ascribe the mental belief to the human mind, to differentiate belief from the physical reality, and even to recognize false-belief and perform psychological reasoning, is a significant milestone in the acquisition of \ac{tom}~\cite{saracho2014theory,wimmer1983beliefs}. Such evidence emerged from developmental psychology in the past few decades call for integrating such socio-cognitive aspects into a modern social robot~\cite{breazeal2009embodied}.

In fact, false-belief is not rare in our daily life. Two examples are depicted in the middle and the right of \cref{fig:sally}: (i) Where does Bob think his cup1 is after Charlie put cup2 (visually identical to cup1) on the table while Dave took cup1 away? (ii) Which milk box should Alice give to Bob if she wants to help? The one closer to Bob but empty, or the one further to Bob but full? Although such false-belief tasks are primal examples for social and cognitive intelligence, current state-of-the-art intelligent systems are still facing challenges in acquiring such a capability in the wild with noisy visual input (see \emph{Related Work} for discussion).

One fundamental challenge is the lack of proper representation for modeling the false-belief from visual input; it has to be able to handle the heterogeneous information of a system's current states, its accumulated knowledge, agent's belief, and the reality/ground-truth of the world. Without a unified representation, the information across all these domains cannot be easily interpreted, and the cross-domain reasoning of the events is infeasible.

Largely due to this difficulty, prior work that takes noisy sensory input can only solve a sub-problem in understanding false-belief. For instance, sensor fusion techniques are mainly used to obtain better state estimation by filtering the measurements from multiple sensors~\cite{liggins2017handbook}. Similarly, the \ac{mvt} in computer vision is designed to combine the observations across camera views to better track an object. Visual cognitive reasoning (\eg, human intention/attention predictions~\cite{koppula2013learning,qi2017predicting,fan2018inferring,wei2018and}) only targets to model human mental states. These three lines of work are all crucial ingredients but developed independently; a unified cross-domain representation is still largely missing.

In order to endow such an ability to understand the concept of false-belief to a robot system from noisy visual inputs, this paper proposes to use a graphical model represented by a \acf{pg}~\cite{zhu2007stochastic} to serve as the unified representation of a robot's knowledge structure, fused knowledge across all robots, and the (false-)beliefs of human agents. A \ac{pg} is learned from the spatiotemporal transition of humans and objects in the scene perceived by a robot. A joint \ac{pg} can be induced by merging and fusing the individual \ac{pg} from each robot to overcome the errors originated from a single view. In particular, our system enables the following three capabilities with increasing depth in cognition:
\begin{enumerate}[leftmargin=*,noitemsep,nolistsep]
    \item \emph{Tracking small objects with occlusions across different views}. Human-made objects in an indoor environment (\eg, cups) are oftentimes small with a similar appearance. Tracking such objects could be challenging due to occlusions with frequent human interactions. The proposed method can address the challenging multi-view multi-object tracking problem by properly maintaining cross-view object states using the unified representation.
    \item \emph{Inferring human beliefs}. The state of an object normally does not change unless a human interacts with it; this observation shares a similar spirit in human cognition known as object permanence~\cite{baillargeon1985object}. By identifying the interactions between humans and objects, our system also supports the high-level cognitive capability; \eg, knowing which object is interacted with which person, whether a person knows the state of the object has been changed.
    \item \emph{Assisting agents by recognizing false-belief}. Giving the above object tracking and cognitive reasoning of human beliefs, the proposed algorithm can further infer whether and why the person has false-belief, thereby to better assist the person given a specific context.
\end{enumerate}

\subsection{Related Work}\label{sec:related_work}

\textbf{Robot \ac{tom}}, aiming at understanding human beliefs and intents, receives increasing research attentions in human-robot interaction and collaboration~\cite{scassellati2002theory,thomaz2016computational}. Several false-belief tasks akin to the classic Sally-Anne test were designed. For instance, Warnier \etal~\cite{warnier2012robot} introduced a belief management algorithm, and the reasoning capability is subsequently endowed to a robot to pass the Sally-Anne test~\cite{milliez2014framework} successfully. More sophisticated human-robot collaboration is achieved by maintaining a human partner's mental state~\cite{devin2016implemented}. More formally, Dynamic Epistemic Logic is introduced to represent and reason about belief and false-belief~\cite{bolander2018seeing,lorini2019decision}. These successes are, however, limited to the symbolic-based belief representations, requiring handcrafted variables and structures, making the logic-based reasoning approaches brittle in practice to handle noises and errors. To address this deficiency, this paper utilizes a unified representation by \ac{pg}, a probabilistic graphical model that has been successfully applied to various robotics tasks, \eg,~\cite{edmonds2017feeling,liu2018interactive,edmonds2019tale}; it accumulates the observations over time to form a knowledge graph and robustly handles noisy visual input.

\textbf{Multi-view Visual Analysis} is widely applied to 3D reconstruction~\cite{hofmann2013hypergraphs}, object detection~\cite{liebelt2010multi,utasi20113}, cross-view tracking~\cite{berclaz2011multiple,xu2016multi}, and joint parsing~\cite{qi2018scene}. Built on top of these modules, \ac{mot} usually utilizes tracking-by-detection techniques~\cite{wen2014multiple,dehghan2015target,dong2017occlusion}. This line of work primarily focuses on combining different camera views to obtain a more comprehensive tracking, lacking the understanding of human (false-)belief.

\textbf{Visual Cognitive Reasoning} is an emerging field in computer vision. Related work includes recovering incomplete trajectories~\cite{liang2018tracking}, learning utility and affordance~\cite{zhu2016inferring}, inferring human intention and attention~\cite{fan2018inferring,wei2018and}, \etc. As to understanding (false-)belief, despite many psychological experiments and theoretical analysis~\cite{call1999nonverbal,baker2009action,brauner2016second,yang2018emotion}, very few attempts have been made to solve (false-)belief with visual input; handcrafted constraints are usually required for specific problems in prior work. In contrast, this paper utilizes a unified representation across different domains with heterogeneous information to model human mental states.

\subsection{Contribution}

This paper makes three contributions:
\begin{enumerate}[leftmargin=*]
    \item We adopt a unified graphical model \ac{pg} to represent and maintain heterogeneous knowledge about object states, robot knowledge, and human beliefs.
    \item On top of the unified representation, we propose an inference algorithm to merge individual \ac{pg} from different domains across time and views into a joint \ac{pg}, supporting human belief inference from multi-view to overcome the noises and errors originated from a single view.
    \item With the inferred \ac{pg}s, our system can keep track of the state and location of each object, infer human beliefs, and further discover false-belief to better assist human.
\end{enumerate}

\subsection{Overview}

The remainder of the paper is organized as follows. \Cref{sec:representation,sec:formulation} describe the representation and the detailed probabilistic formulation, respectively. We demonstrate the efficacy of the proposed method in \cref{sec:experiment} and conclude the paper with discussions in \cref{sec:conclusion}.

\begin{figure*}[t!]
    \centering
    \includegraphics[width=\linewidth,trim={3cm 4cm 3cm 4cm},clip]{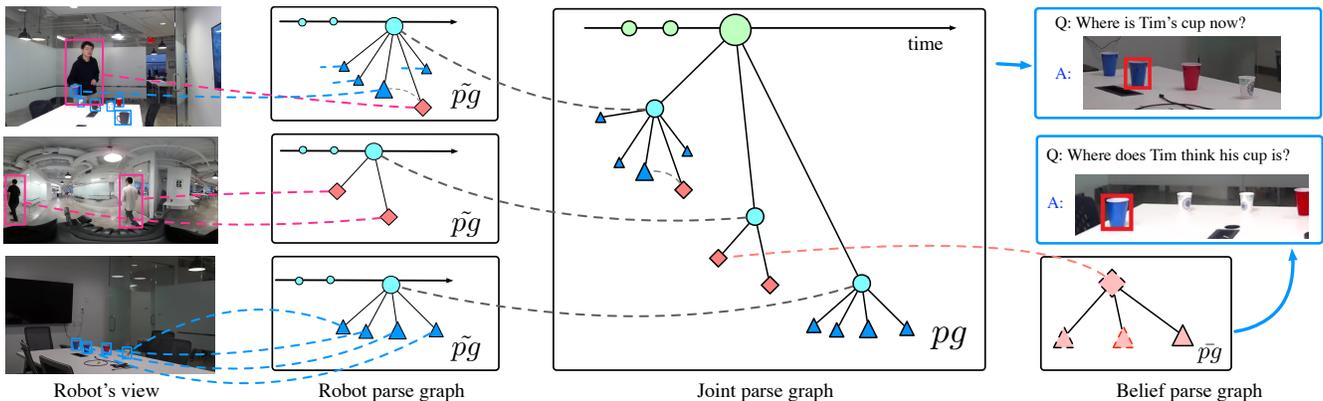}
    \caption{System overview. The robot \ac{pg}s are obtained from each individual robot's view. The joint \ac{pg} can be obtained by fusing all robots' \ac{pg}s. The belief \ac{pg}s can be inferred from the joint \ac{pg}. All the \ac{pg}s are optimized simultaneously under the proposed joint parsing framework to enable the queries about the object states and human (false-)beliefs.}
    \label{fig:pipeline}
\end{figure*}

\section{Representation}\label{sec:representation}

In this work, we use the \acf{pg}---a unified graphical model~\cite{zhu2007stochastic}---to represent (i) the location of each agent and object, (ii) the interactions between agents and objects, (iii) the beliefs of agents, and (iv) the attributes and states of objects; see \cref{fig:pipeline} for an example. Specifically, three different types of \ac{pg}s are utilized:
\begin{itemize}[leftmargin=*,noitemsep,nolistsep]
    \item \emph{Robot \ac{pg}}, shown as blue circles, maintains the knowledge structure of an individual robot, which is extracted from its visual observation---an image. It also contains attributes that are grounded to the observed agents and objects.
    \item \emph{Belief \ac{pg}}, shown as red diamonds, represents the \emph{inferred} human knowledge by each robot. Each robot maintains the parse graph for each agent it observed.
    \item \emph{Joint \ac{pg}} fuses all the information and views across a set of distributed robots.
\end{itemize}

\paragraph*{Notations and Definitions}

The input of our system can be represented by $M$ synchronized video sequences $I = \{I_{t=1..T}^{k=1..M}\}$ with length $T$ captured from $M$ robots. Formally, a scene $\mathcal{R}$ is expressed as
\begin{equation}\begin{aligned}
    \mathcal{R} &= \{(O_t, H_t): t = 1,2,\ldots,T\}, \\
    O_t &= \{o_t^i: i = 1,2,\ldots,N_o\}, \\
    H_t &= \{h_t^j: j = 1,2,\ldots,N_h\},
\end{aligned}\end{equation}
where $O_t$ and $H_t$ denote the set of all the tracked objects ($N_o$ objects in total) and the set of all the tracked agents ($N_h$ agents in total) at time $t$, respectively.

Object $o^i_t$ is represented by a tuple: bounding box location $b^i_t$, appearance feature $\phi^i_t$, states $s^i_t$, and attributes $a^i_t$,
\begin{equation}
    o_t^i = (b^i_t, \phi_t^i, s_t^i, a_t^i),
\end{equation}
where $s^i_t$ is an index function: $s^i_t = j, j \neq 0$ indicates the object $o_i$ is held by the agent $h_j$ at time $t$, and $s^i_t = 0$ means it is not held by any agent at time $t$.

The agent $h_t^j$ is represented by its body key-point position $\kappa^j_t$ and appearance feature $\phi_t^j$
\begin{equation}
    h_t^j = (\kappa^j_t, \phi_t^j).
\end{equation}

\emph{Robot Parse Graph} is formally expressed as
\begin{equation}
    \tilde{pg}^k_t = \{(o_t^i, h_t^j): o_t^i, h_t^j \in I^k_t \},
\end{equation}
where $I^k_t$ is the area where $k$th robot can observe at time $t$.

\emph{Belief Parse Graph} is formally expressed as
\begin{equation}
    \bar{pg}^{k,j}_t = \{o_{t'}^i: o_{t'}^i \in I^k_{t'} \},
    \label{eqn:mindpg}
\end{equation}
where $\bar{pg}^{k,j}_t$ represents the inferred belief of agent $h_j$ under robot $k$'s view; $t'$ is the last time that the robot $k$ observes the human $h_j$. We assume that the agent $h_j$ only keeps the objects s/he observed last time in this area in mind, which satisfies the \emph{Principle of Inertia}: an agent's belief is preserved over time unless the agent gets information to the contrary.

\emph{Joint Parse Graph} keeps track of all the information across a set of distributed robots, formally expressed as
\begin{equation}
    pg_t = \{(o_t^i, h_t^j: i = 1,2,...,N_o; j = 1,2,...,N_h) \}. \\
\end{equation}

\paragraph*{Objective}

The objective of the system is to jointly infer all the parse graphs $PG = \{pg, \tilde{pg}, \bar{pg}\}$ so that it can (i) track all the agents and objects across scenes at any time by fusing the information collected by a distributed system, and (ii) infer human (false-)beliefs to provide assistance.

\section{Probabilistic Formulation}\label{sec:formulation}

We formulate the joint parsing problem as a \ac{map} inference problem
\begin{fleqn}\begin{equation}
\resizebox{0.93\hsize}{!}{$\displaystyle%
    PG^* = \argmax_{PG} p(PG | I) = \argmax_{PG} p(I | PG) \cdot p(PG),
$}%
\end{equation}\end{fleqn}
where $p(PG)$ is the prior, and $p(I | PG)$ is the likelihood.

\subsection{Prior}

The prior term $p(PG)$ models the compatibility of the robot \ac{pg}s and the joint \ac{pg}, and the compatibility of the joint \ac{pg} over time. Formally, we can decompose the prior as
\begin{equation}
    p(PG) = p(pg_1)\prod_{t=1}^{T-1} p(pg_{t+1} | pg_{t}) \,\prod_{k=1}^{M} \prod_{t=1}^{T} p(\tilde{pg}_t^{k} | pg_t),
\end{equation}
where the first term $p(pg_{t+1} | pg_{t})$ is the transition probability of the joint \ac{pg} over time, further decomposed as
\begin{equation}
    p(pg_{t+1} | pg_{t}) = \dfrac{1}{Z} \exp \{ - \energy(pg_{t+1} | pg_{t}) \},
\end{equation}
\begin{equation}
\resizebox{0.89\hsize}{!}{$\displaystyle%
    \energy(pg_{t+1} | pg_{t}) = \sum_{i=1}^{N_o} \energy_{L_o}(b^i_{t+1}, b^i_{t}, s_t^i) + \energy_{ST}(s_{t+1}^i, s_t^i) + \sum_{j=1}^{N_h} \energy_{L_h}(\kappa^j_{t+1}, \kappa^j_{t}).
    \label{eq:prior1}
$}%
\end{equation}

The second term $p(\tilde{pg}_t^{k} | pg_t)$ is the probability which models the compatibility of individual \ac{pg}s and the joint \ac{pg}. Its energy can be decomposed into three energy terms
\begin{fleqn}\begin{equation*}
    p(\tilde{pg}_t^k | pg_t) = \dfrac{1}{Z} \exp \{ - \energy(pg_t, \tilde{pg}_t^k) \}
\end{equation*}\end{fleqn}
\begin{equation}
\resizebox{0.89\hsize}{!}{$\displaystyle%
    = \dfrac{1}{Z} \exp \{ - \energy_A(pg_t, \tilde{pg}_t^k) - \energy_S(pg_t, \tilde{pg}_t^k) - \energy_{Attr}(pg_t, \tilde{pg}_t^k) \}.
    \label{eq:prior2}
$}%
\end{equation}

Below, we detail the above six energy terms $\energy_{(\cdot)}$ in \cref{eq:prior1,eq:prior2}.

\paragraph*{Motion Consistency}

The term $\energy_{L}$ measures the motion consistency of objects and agents in time, defined as
\begin{equation}\begin{aligned}
    \energy_{L_o}(b^i_{t+1}, b^i_{t}, s^i_t) &=
    \begin{cases}
    \delta(\mathcal{D}(b^i_{t+1}, b^i_{t}) > \tau)) & \text{if $s_t^i = 0$}\\
    \delta(\mathcal{D}(\kappa^j_{t+1}, \kappa^j_{t}) > \tau)) & \text{if $s_t^i = j$}\\
    \end{cases}\\
    \energy_{L_h}(\kappa^j_{t+1}, \kappa^j_{t}) &= \delta(\mathcal{D}(\kappa^j_{t+1}, \kappa^j_{t}) > \tau)),
\end{aligned}\end{equation}
where $\mathcal{D}$ is the distance between two bounding boxes or human poses, $\tau$ is the speed threshold, and $\delta$ is the indicator function. If an object $i$ is held by an agent $j$, we use the agent's location to calculate $\energy_L$ of the object.

\paragraph*{State Transition Consistency}

The term $\energy_{ST}$ is the state transition energy, defined as
\begin{equation}
    \energy_{ST}(s_{t+1}^i, s_t^i) = -\log p(\delta(s_{t+1}^i = 0) | \delta(s_t^i = 0)),
\end{equation}
where the state transition probability $p(\delta(s_{t+1}^i = 0) | \delta(s_t^i = 0))$ is learned from the training data.

\begin{figure*}[t!]
    \centering
    \begin{subfigure}[b]{0.333\linewidth}
        \includegraphics[width=\linewidth]{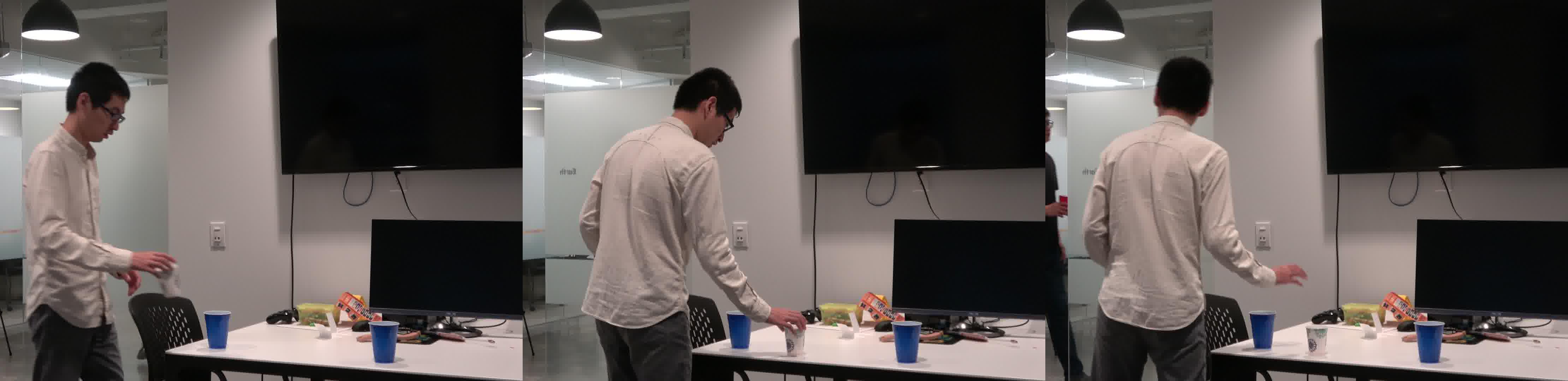}
        \caption{Put down a cup}
    \end{subfigure}%
    \begin{subfigure}[b]{0.333\linewidth}
        \includegraphics[width=\linewidth]{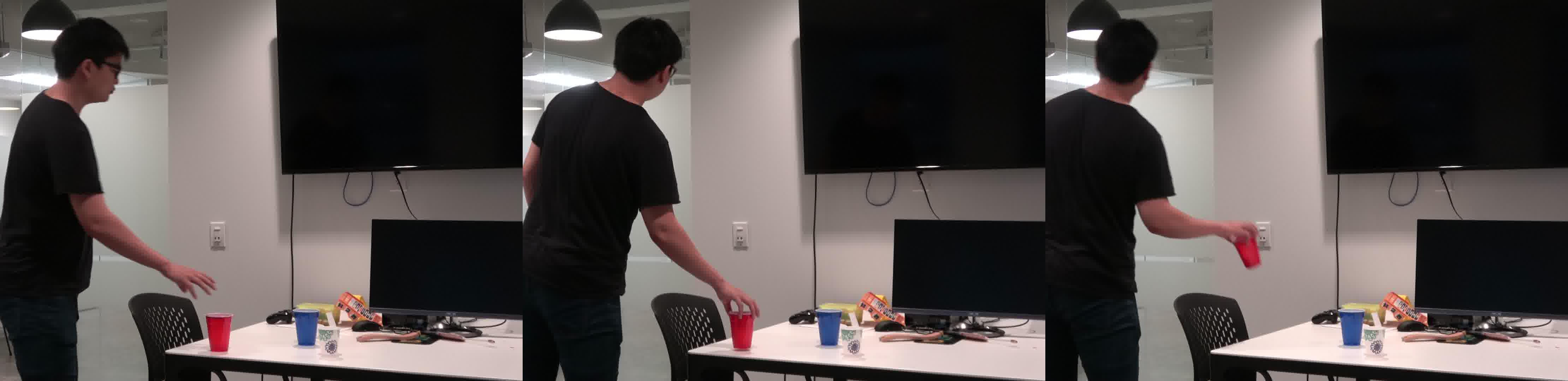}
        \caption{Pick up a cup}
    \end{subfigure}%
    \begin{subfigure}[b]{0.333\linewidth}
        \includegraphics[width=\linewidth]{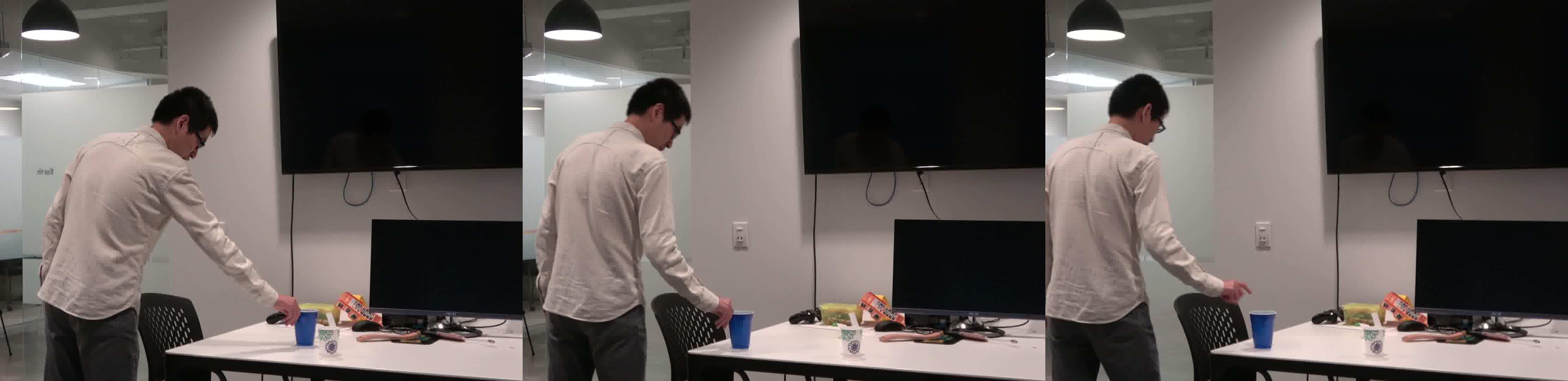}
        \caption{Move a cup}
    \end{subfigure}%
    \\
    \begin{subfigure}[b]{0.555\linewidth}
        \includegraphics[width=\linewidth]{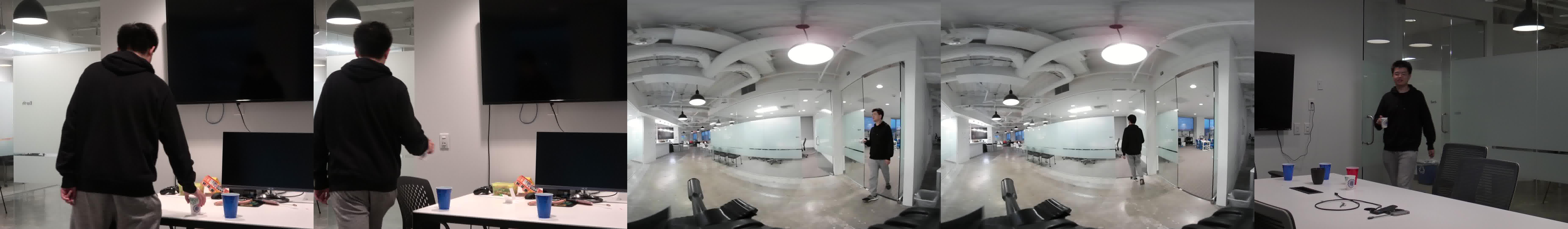}
        \caption{Carry a cup to another room}
    \end{subfigure}%
    \begin{subfigure}[b]{0.445\linewidth}
        \includegraphics[width=\linewidth]{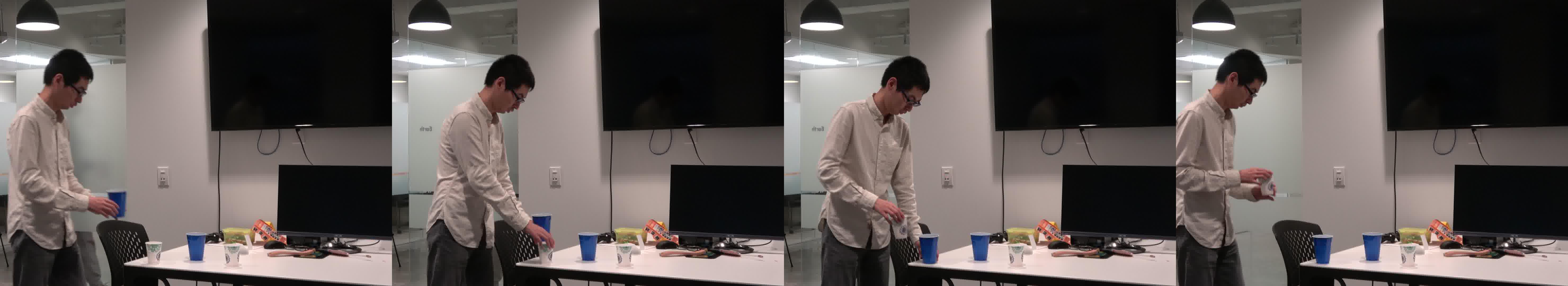}
        \caption{Swap two cups}
    \end{subfigure}%
    \caption{Examples of human-object interactions in the cross-view subset of the proposed dataset. Each scenario contains at least one kind of false-belief test or helping test recorded with four robot camera views.}
    \label{fig:interaction}
\end{figure*}

\paragraph*{Appearance Consistency}

$\energy_A$ measures appearance consistency. In robot \ac{pg}s, the appearance feature vector $\phi$ is extracted by a deep person re-identification network~\cite{zhou2019omni}. In the joint \ac{pg}, the feature vector is calculated by the mean pooling of all features for the same entity from all robot \ac{pg}s
\begin{equation}
    \energy_A(pg_t, \tilde{pg}_t) = \sum_{e \in O_t \cup H_t} || \phi^e_t - \phi^{\tilde{e}}_t ||_2.
\end{equation}

\paragraph*{Spatial Consistency}

Each object and agent in the robot's 2D view should also have a corresponding 3D location in the real-world coordinate system, and such correspondence should remain consistent when projected any points from the robot image plane back to the real-world coordinate. Thus, spatial consistency is defined as
\begin{equation}
    \energy_S(pg_t, \tilde{pg}_t) = \sum_{e \in O_t \cup H_t} || \Lambda^e_t - f(\Lambda^{\tilde{e}}_t)||_2,
\end{equation}
where $\Lambda$ is the 3D positions in the real-world coordinate, and $f$ is the transformation function that projects the points from the robot's 2D view to the 3D real-world coordinate.

\paragraph*{Attribute Consistency}

Attributes of each entity should remain the same across time and viewpoints. Such an attribute consistency is defined by the term $\energy_{Attr}$
\begin{equation}
    \energy_{Attr}(pg_t, \tilde{pg}_t) = \sum_{i = 1}^{N_o} \delta(a^i_t\neq\tilde{a}^i_t).
\end{equation}

\subsection{Likelihood}

The likelihood term $p(I | PG)$ models how well each robot can ground the knowledge in its \ac{pg} to the visual data it captures. Formally, the likelihood is defined as
\begin{equation}
    p(I | PG) = \prod_{k=1}^{M} \prod_{t=1}^{T} p(I_t^{k} | \tilde{pg}_{t}^{k}).
\end{equation}
The energy of term $p(I_t^{k} | \tilde{pg}_{t}^{k})$ can be further decomposed as
\begin{equation}
    p(I_t^{k} | \tilde{pg}_{t}^{k}) = \dfrac{1}{Z} \exp \{ - \energy(I_t^{k} | \tilde{pg}_{t}^{k}) \},
\end{equation}
\begin{fleqn}\begin{equation}
\resizebox{0.91\hsize}{!}{$\displaystyle%
    \energy(I_t^{k} | \tilde{pg}_{t}^{k}) =\sum_{i=1}^{N_o} \energy_{D}(b^i_{t}, \phi_t^i) + \energy_{C}(b^i_{t}, \phi_t^i, a^i_t) + \sum_{j=1}^{N_h} \energy_{D}(p^j_{t}, \phi^j_{t}),
$}%
\end{equation}\end{fleqn}
where $\energy_{D}$ can be calculated by the score of object detection or human pose estimation, and $\energy_{C}$ can be obtained by the object attributes classification scores.

\subsection{Inference}

Given the above probabilistic formulation, we can infer the best $\{pg^*, \tilde{pg}^*\}$ by an \ac{map} estimate. It can be solved by two steps: (i) Each robot individually processes the visual input; the output (\eg, object detection, and human pose estimation) can be aggregated as the proposals for the second step. (ii) The \ac{map} estimate can be transformed to an assignment problem given the proposals, solvable using the \emph{Kuhn-Munkres} algorithm~\cite{kuhn1955hungarian,kuhn1956variants} in polynomial time.

Based on \cref{eqn:mindpg}, robot $k$ can generate belief parse graphs $\bar{pg}^{k,j}$ for agent $j$ after obtaining the robot graphs $\tilde{pg}^k$.

\section{Experiment}\label{sec:experiment}

We evaluate the proposed method in two setups: cross-view object tracking and human (false-)belief understanding. The first experiment evaluates the accuracy of object localization using the proposed inference algorithms, focusing on the robot parse graphs $\tilde{pg}$ and the joint parse graph $pg$. The second experiment evaluates the inference of the belief parse graphs $\bar{pg}$, \ie, human beliefs regarding the object states (\eg, locations) in both single-view and multi-view settings.

\subsection{Dataset}

The dataset includes two subsets, a multi-view subset and a single-view subset. Ground-truth tracking results of objects and agents, and states and attributes of objects are all annotated for evaluation purpose.
\begin{itemize}[leftmargin=*,noitemsep,nolistsep]
    \item The single-view subset includes 5 different false-belief scenarios with $12532$ frames. Each scenario contains at least one kind of false-belief test or helping test. In this subset, objects are not limited to the cups.
    \item The multi-view subset consists of 8 scenes, each shot with 4 robot camera views, making a total number of $72720$ frames. Each scenario contains at least one kind of false-belief test. The objects in each scene are, however, limited to the cups: 12-16 different cups made with 3 different materials (plastic, paper, and ceramic) and 4 colors (red, blue, white, and black). In each scene, three agents interact with cups by performing actions depicted in \cref{fig:interaction}.
\end{itemize}

\subsection{Implementation Details}

Below, we detail the implementations of the system.
\begin{itemize}[leftmargin=*,noitemsep,nolistsep]
    \item Object detection: we use the RetinaNet model~\cite{lin2017focal} pre-trained on the MS COCO dataset~\cite{lin2014microsoft}. We keep all the bounding boxes with a score higher than the threshold $0.2$, which serve as the proposals for object detection.
    \item Human pose estimation: we apply the AlphaPose~\cite{fang2017rmpe}.
    \item Object attribute classification: A VGG16 network~\cite{simonyan2014very} was trained to classify the color and the material of the objects.
    \item Appearance feature: A deep person re-id model~\cite{zhou2019omni} was fine-tuned on the training set.
    \item Due to the lack of multi-view in the single-view setting, we locate the object that an agent plan to interact by simply finding the object closest to the direction the agent points at according to the key points on the arm.
\end{itemize}

\setstretch{0.91}

\begin{figure*}[t!]
    \centering
    \begin{subfigure}[b]{0.2\linewidth}
        \begin{overpic}[width=\linewidth]{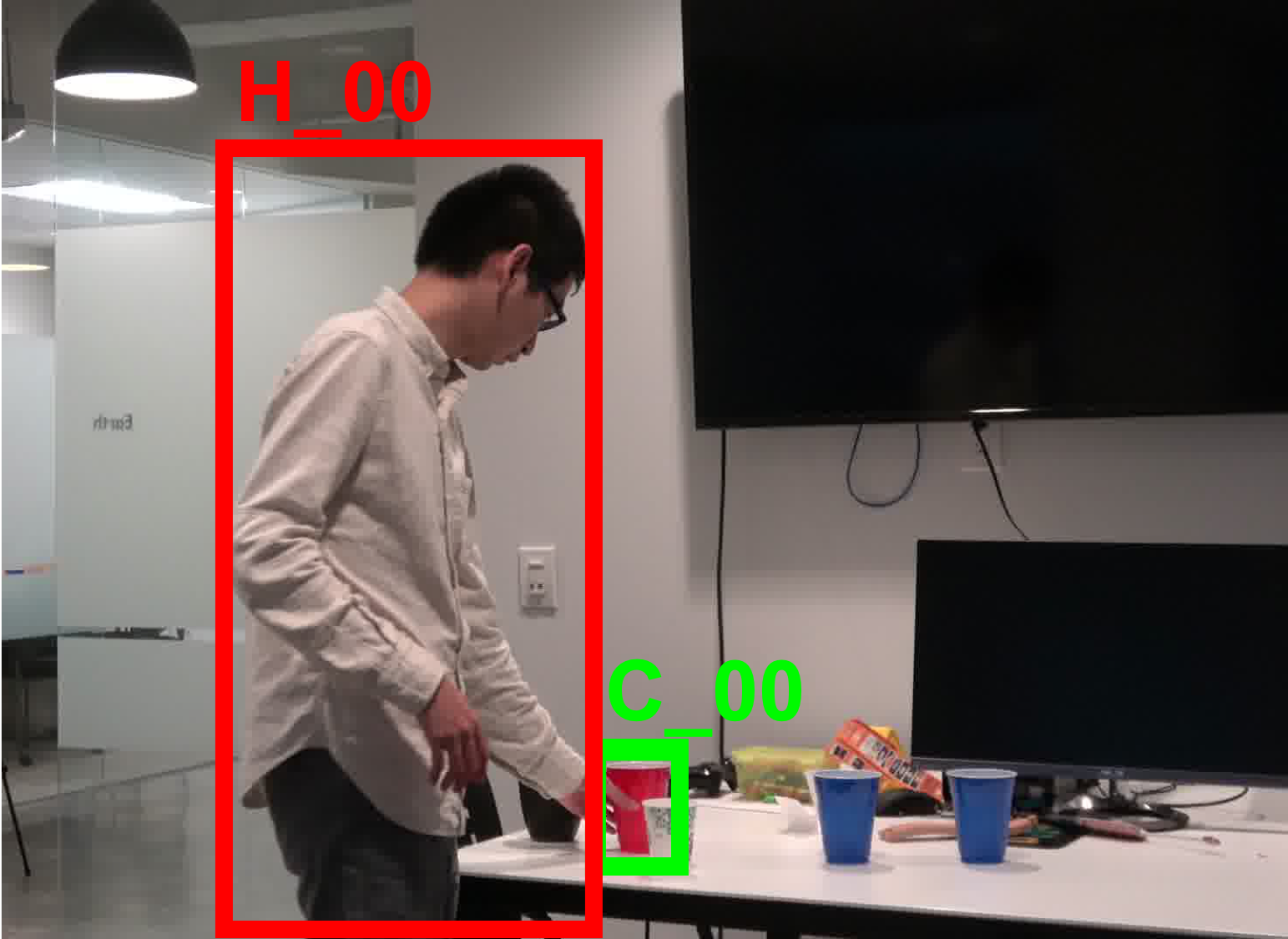}
            \put(65, 55){\color{white}Room 1}
        \end{overpic}
    \end{subfigure}%
    \begin{subfigure}[b]{0.2\linewidth}
        \includegraphics[width=\linewidth]{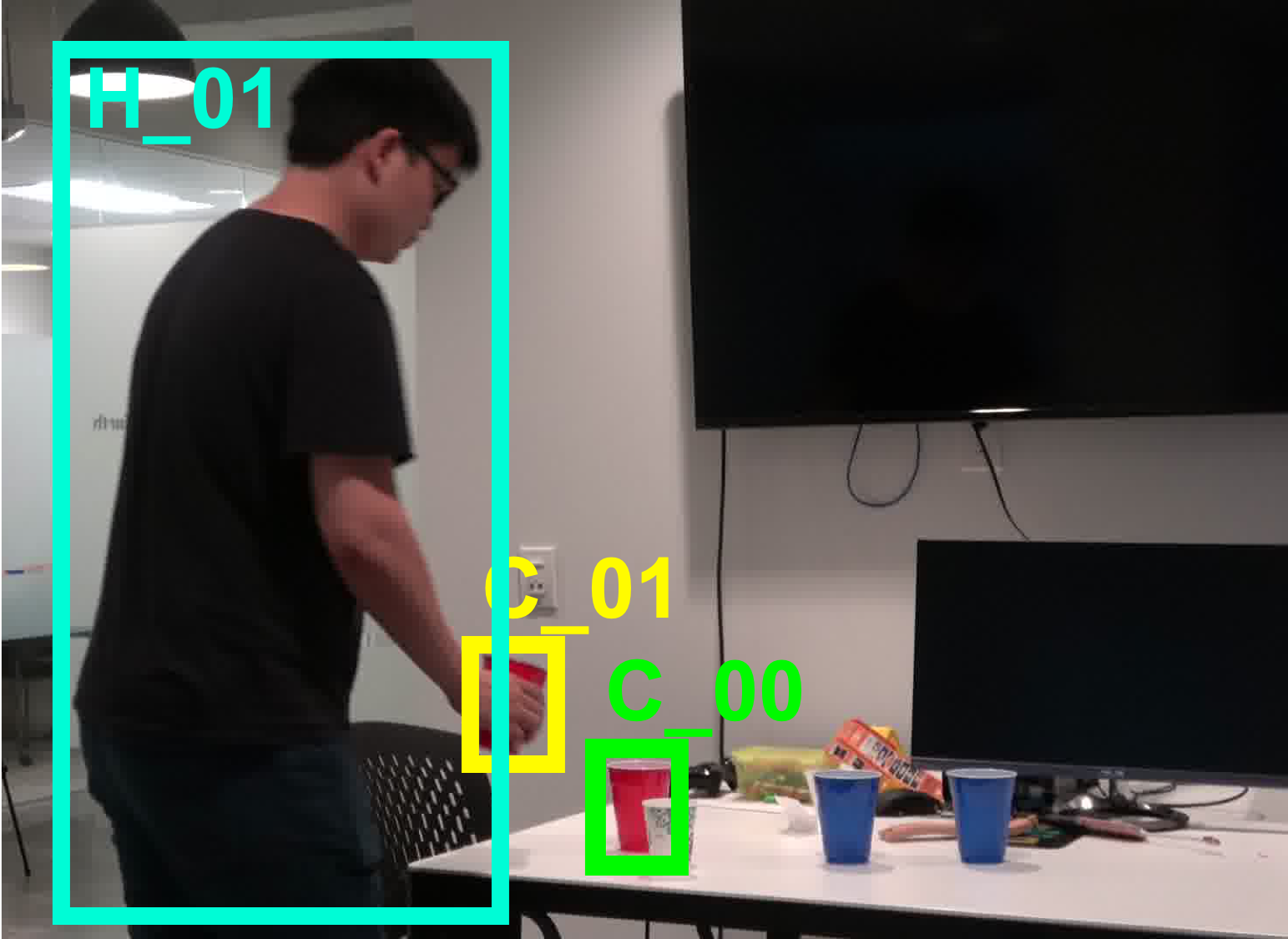}
    \end{subfigure}%
    \begin{subfigure}[b]{0.2\linewidth}
        \includegraphics[width=\linewidth]{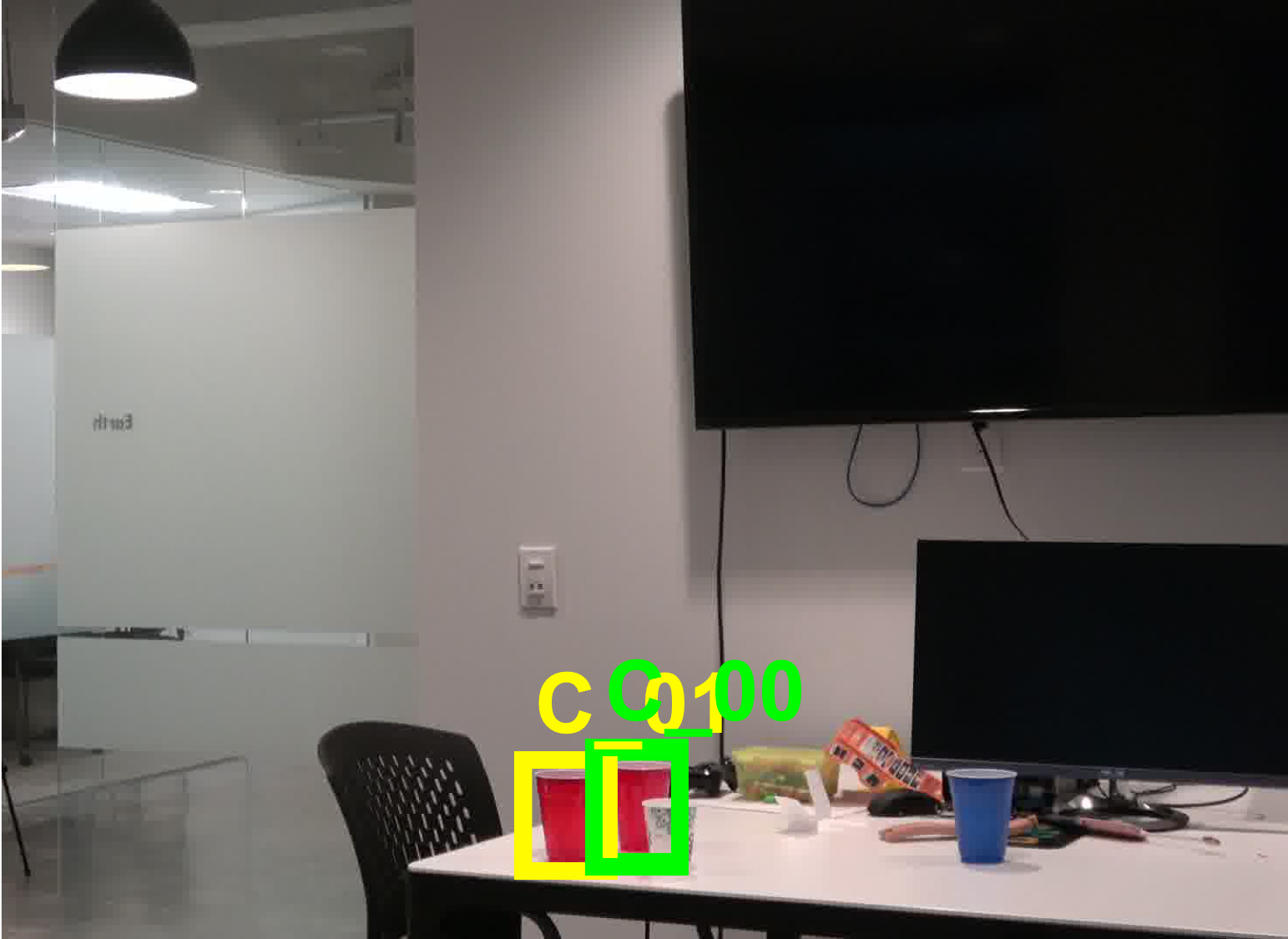}
    \end{subfigure}%
    \begin{subfigure}[b]{0.2\linewidth}
        \includegraphics[width=\linewidth]{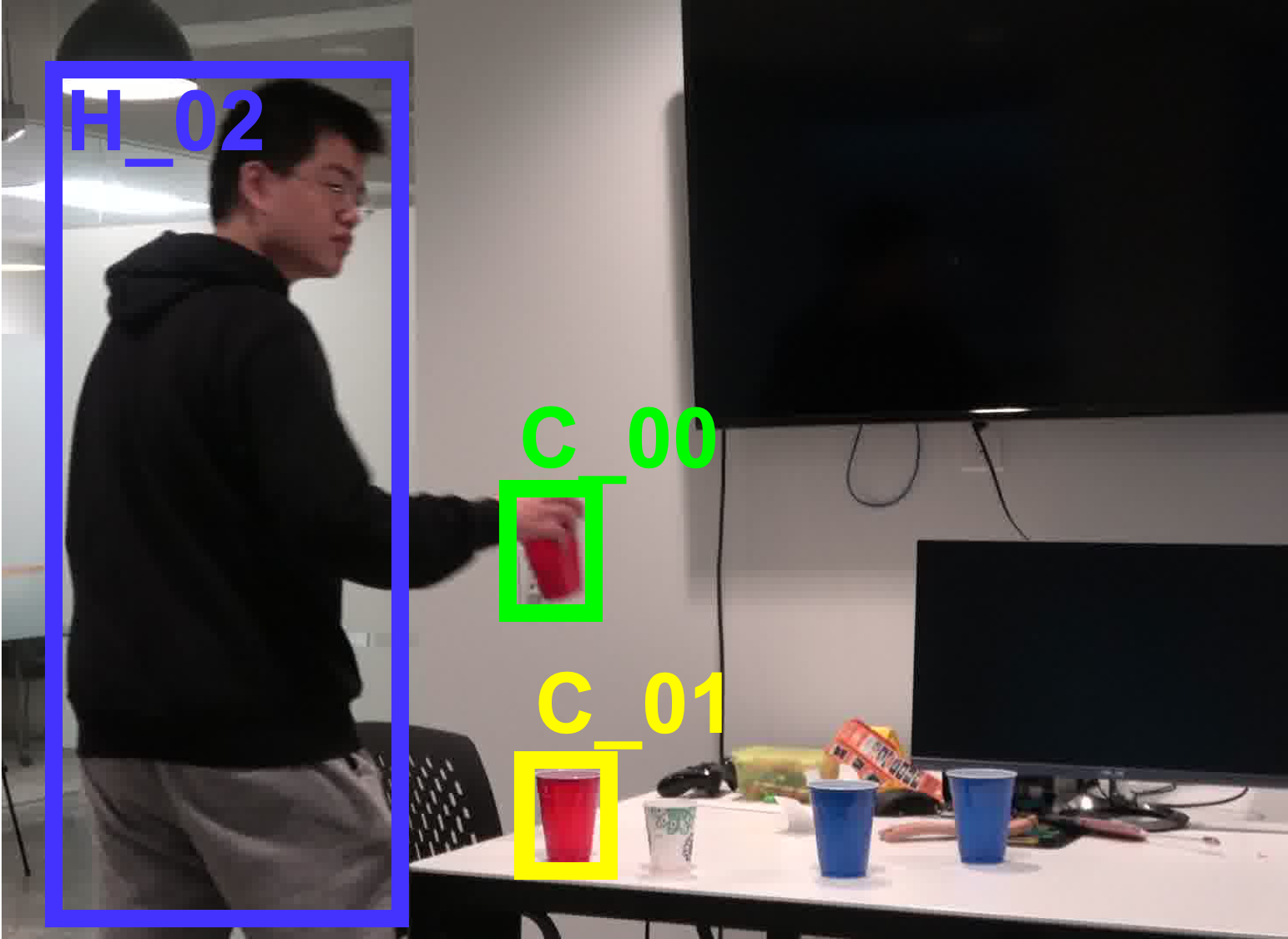}
    \end{subfigure}%
    \begin{subfigure}[b]{0.2\linewidth}
        \includegraphics[width=\linewidth]{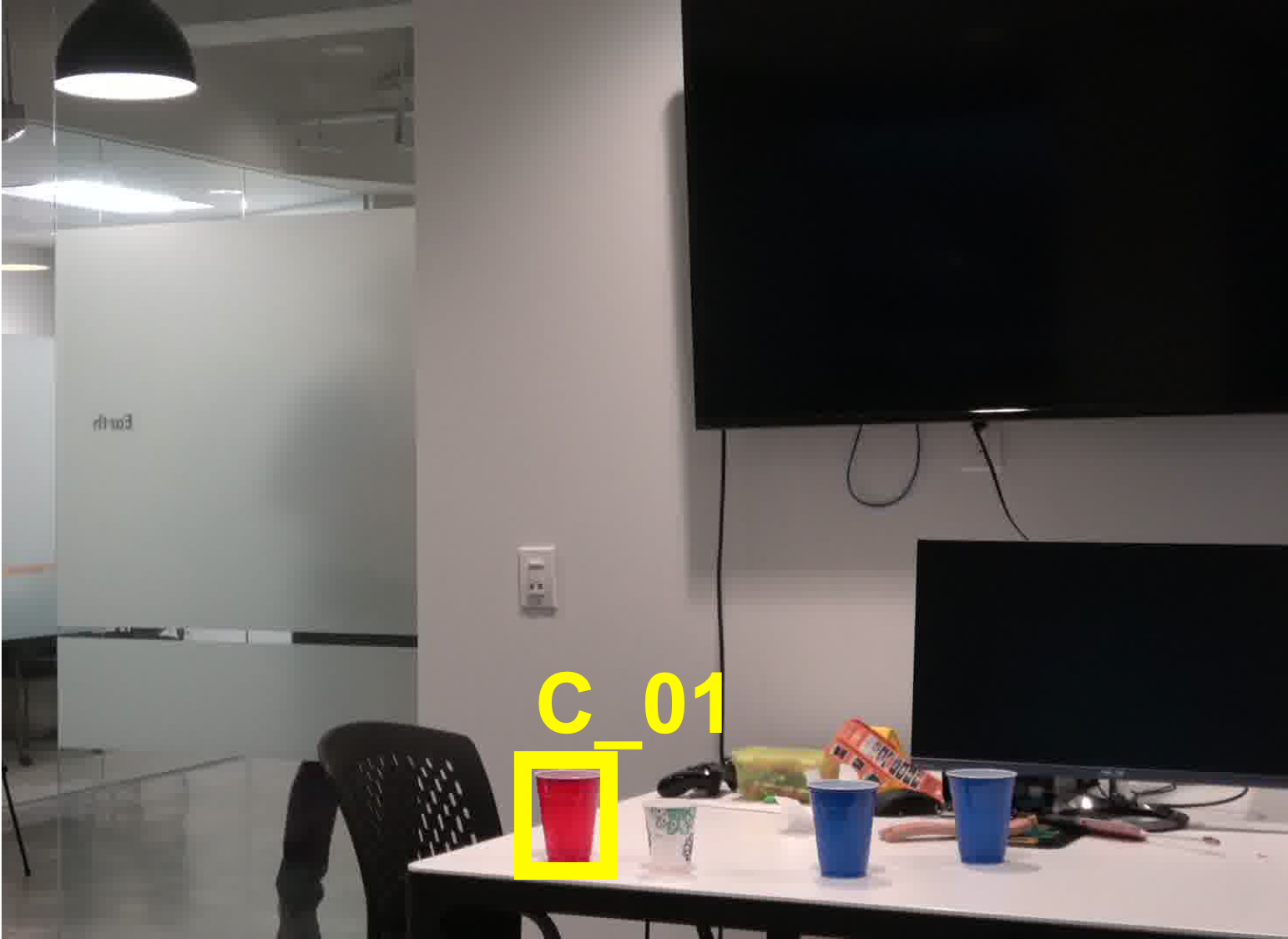}
    \end{subfigure}%
    \\
    \begin{subfigure}[b]{0.2\linewidth}
        \begin{overpic}[width=\linewidth]{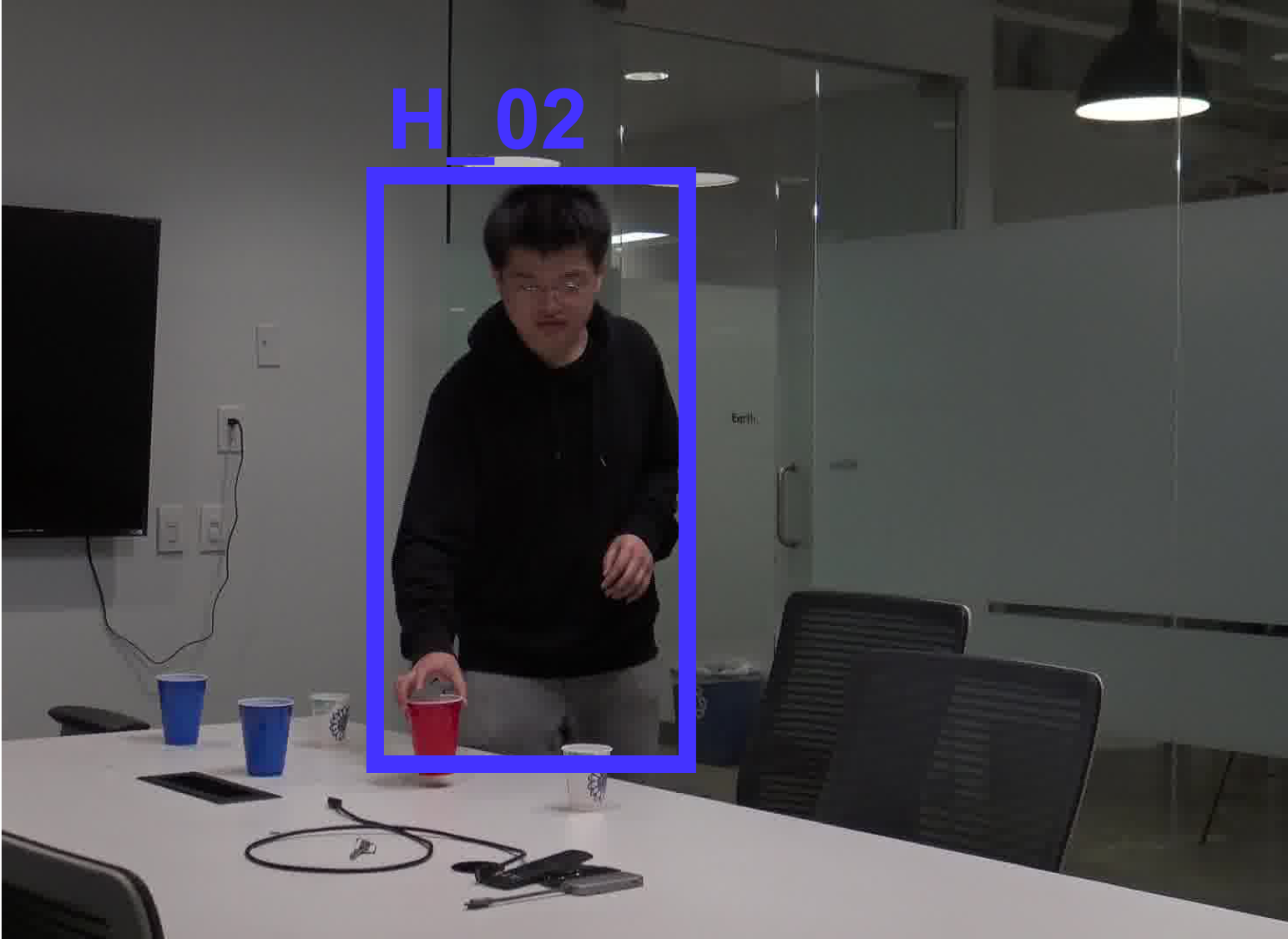}
            \put(65, 45){\color{white}Room 2}
        \end{overpic}
        \caption{Observation $t_0$}
    \end{subfigure}%
    \begin{subfigure}[b]{0.2\linewidth}
        \includegraphics[width=\linewidth]{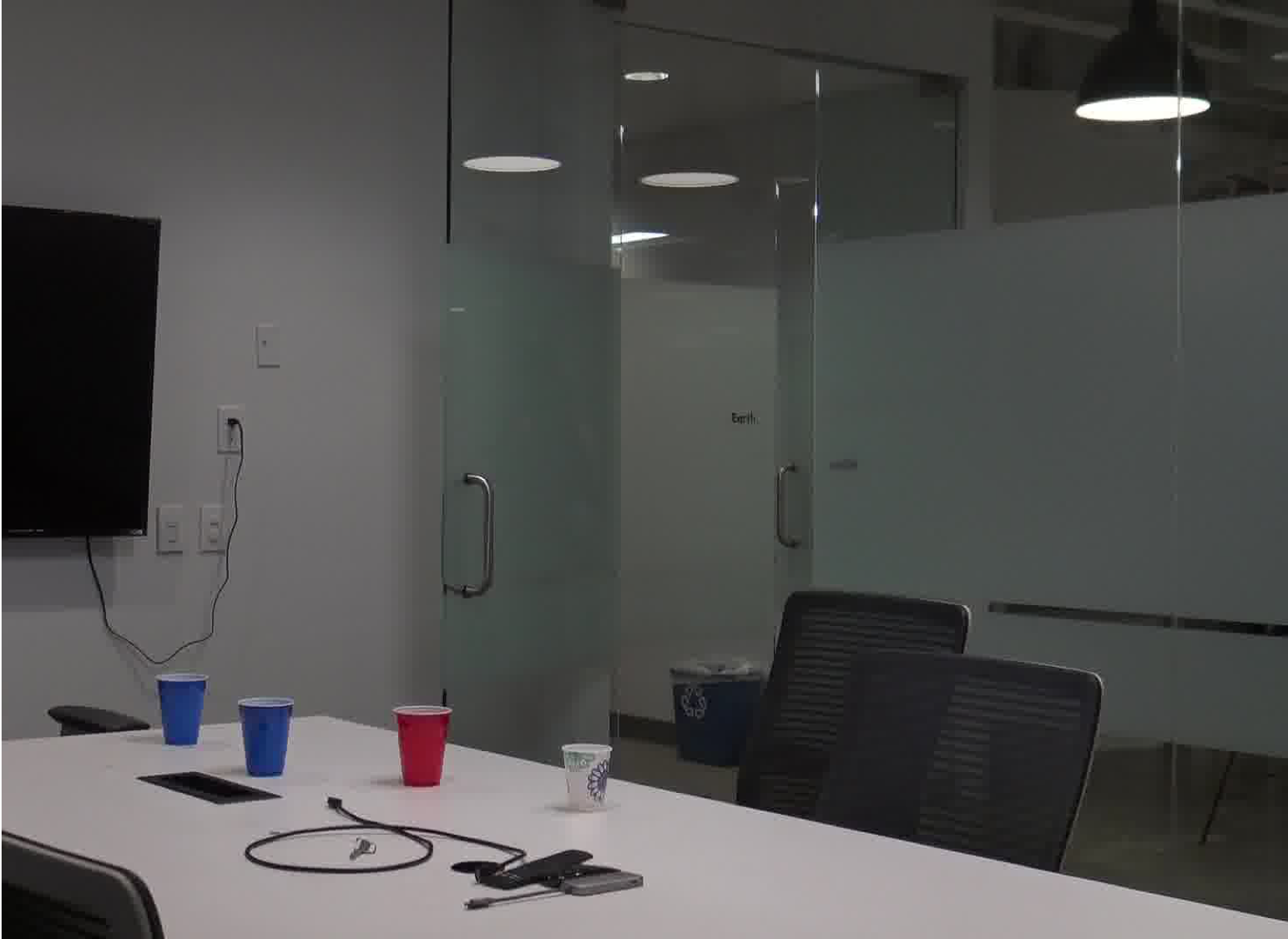}
        \caption{Observation $t_1$}
    \end{subfigure}%
    \begin{subfigure}[b]{0.2\linewidth}
        \includegraphics[width=\linewidth]{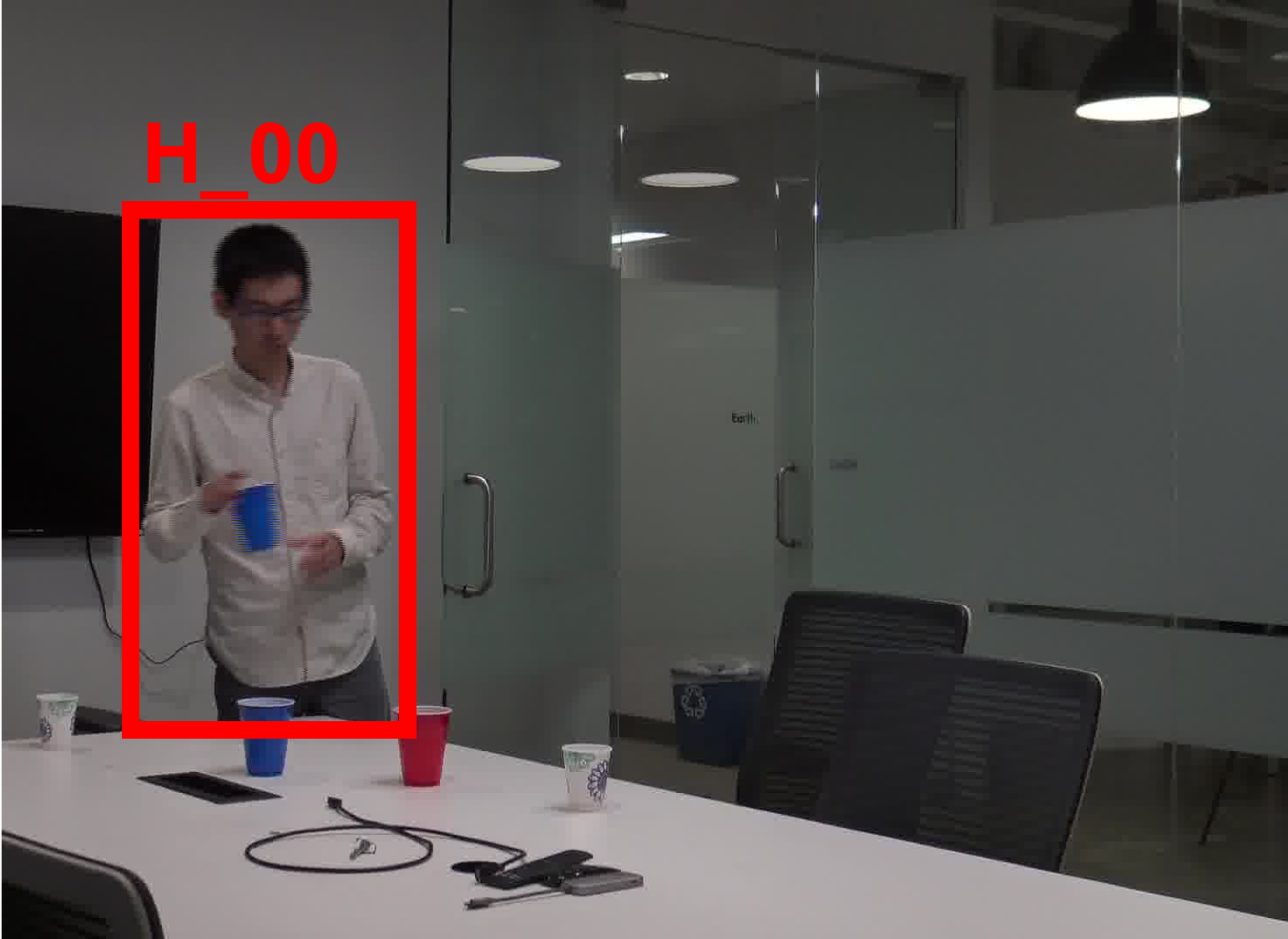}
        \caption{Observation $t_2$}
    \end{subfigure}%
    \begin{subfigure}[b]{0.2\linewidth}
        \includegraphics[width=\linewidth]{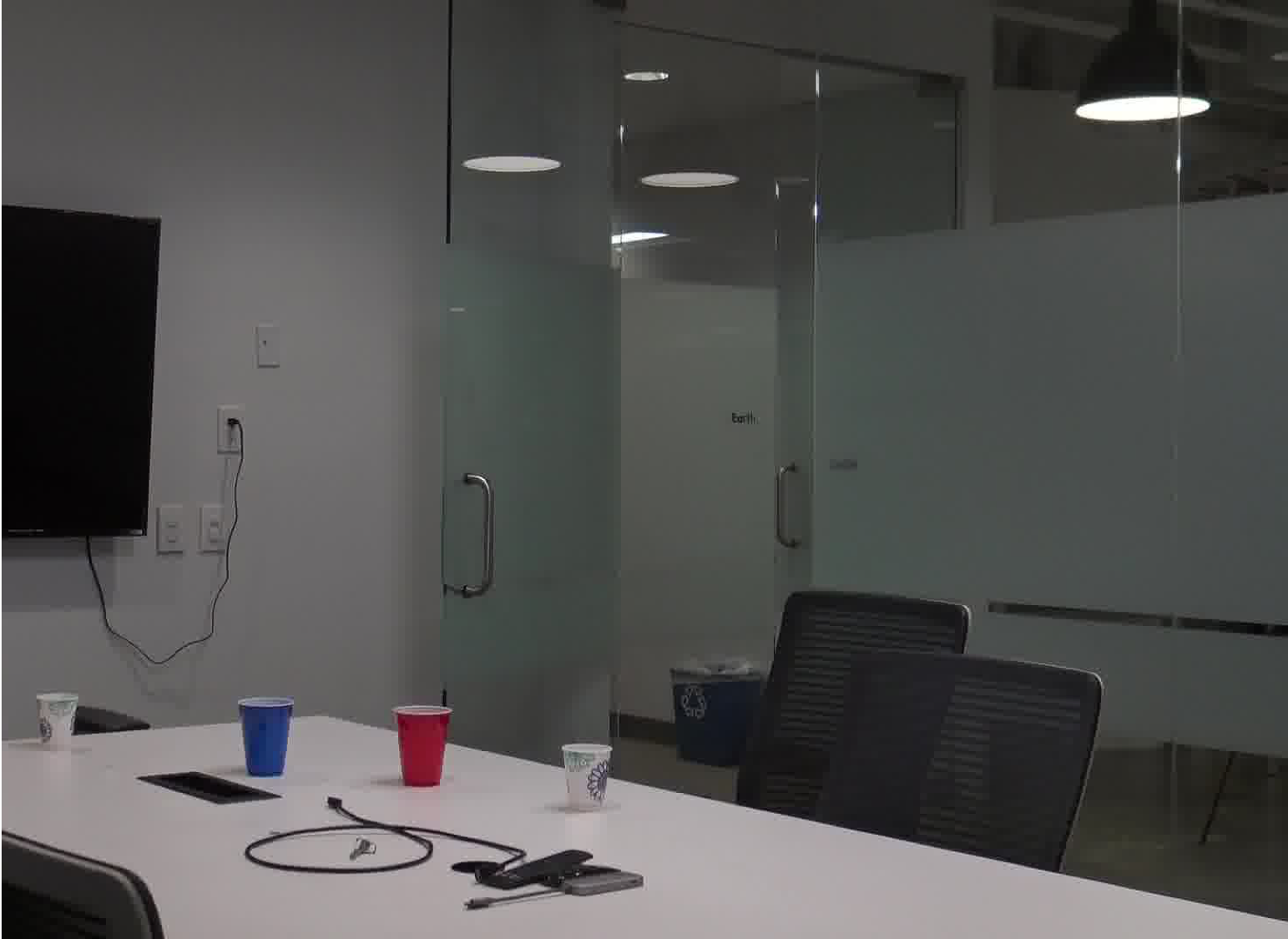}
        \caption{Observation $t_3$}
    \end{subfigure}%
    \begin{subfigure}[b]{0.2\linewidth}
        \includegraphics[width=\linewidth]{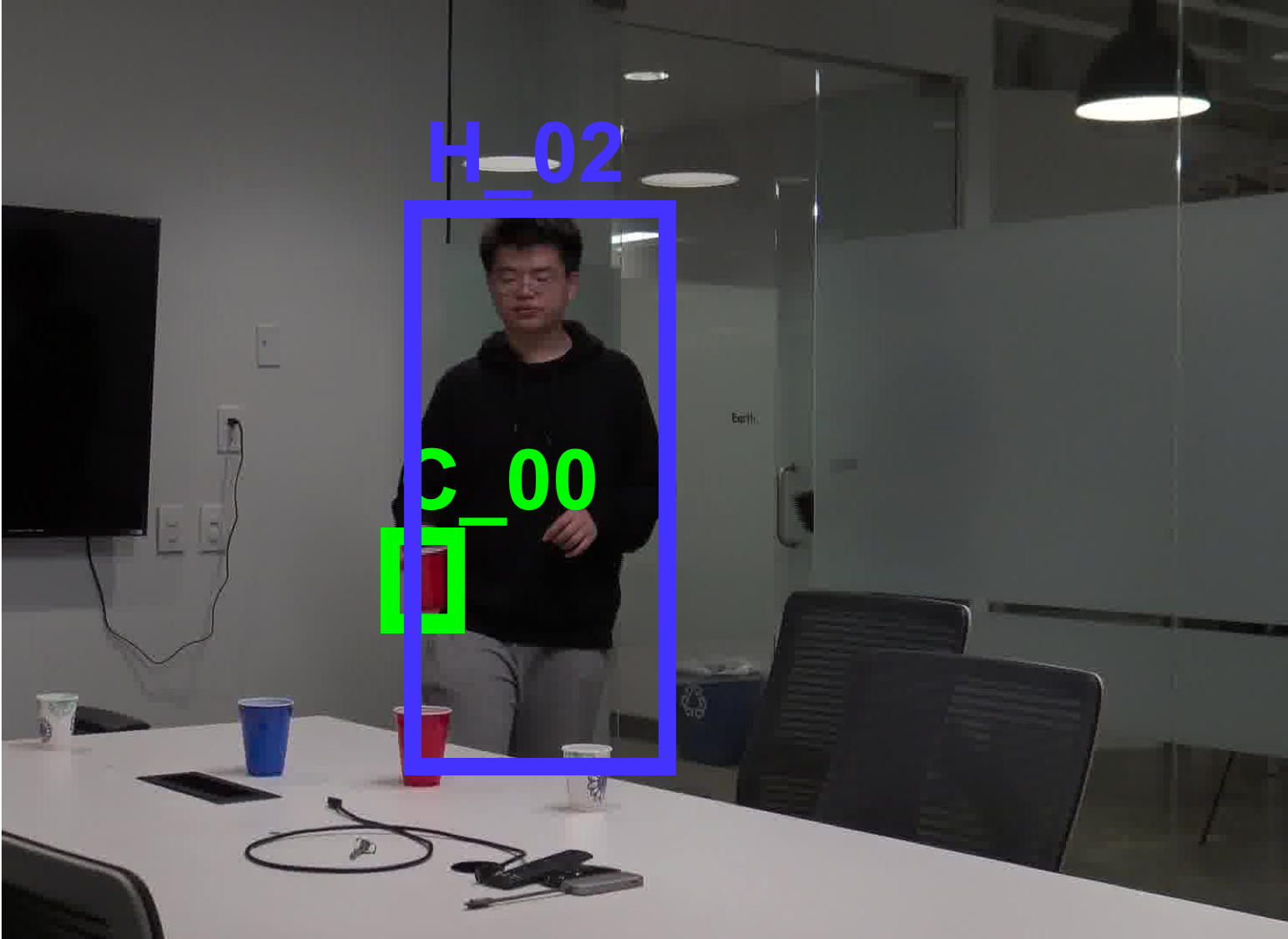}
        \caption{Observation $t_4$}
    \end{subfigure}%
    \caption{A sample result in Experiment 1. Two rows show examples of synchronized visual inputs in different rooms. Two red cups marked by green and yellow bounding boxes are highlighted. At time $t_0$, agent $H_{00}$ put a red cup $C_{00}$ in Room 1. At $t_1$, agent $H_{01}$ put another red cup $C_{01}$ in Room 1. At $t_3$, agent $H_{02}$ took away cup $C_{00}$ and put it in Room 2 at $t_4$. Our system can robustly perform such a complex multi-room multi-view tracking and reason about agent's belief. For instance, at $t_4$, the tracking system knows that the cup $C_{00}$ appeared at $t_0$ is in fact in Room 2 and inferred that human $H_{00}$ thinks the cup is still in Room 1.}
    \label{fig:cup_false_belief}
\end{figure*}

\begin{table}[b!]
    \footnotesize
    \centering
    \caption{Accuracy of cross-view object tracking}\label{tab:cup_loc}
    \begin{tabular}{c|c c c c c}
         $\#$ interactions & 0 & 1 & 2 & 3 & Overall \\ \hline
         Parsing w/o humans acc. & 0.98 & 0.82 & 0.78 & 0.75 & 0.82 \\
         Joint parsing acc. & 0.98 & 0.86 & 0.85 & 0.82 & 0.88
    \end{tabular}
\end{table}

\subsection{Experiment 1: Cross-view Object Localization}

To test the overall cross-view tracking performance, 2000 queries are randomly sampled from the ground-truth tracks. Each query $q$ can be formally described as
\begin{equation}
    q = (k, t, b, t_q),
\end{equation}
where the tuple $(k, t, b)$ indicates the object shown in robot $k$'s view located in bounding box $b$ at time $t$. Such a form of the query can be very flexible. For instance, if we ask about the location of that object at time $t_q$, the system should return an answer in the form of $(k_a, b_a)$, meaning that the system predicts the object is shown in robot $k_a$'s view at $b_a$.

The system generates the answer in two steps. It firstly locates the query of the object by searching the object $i$ in $\tilde{pg}^k_t$ with the smallest distance to the bounding box $b$. Then it returns the location $b_{t_q}^i$ from $\tilde{pg}^{k'}_{t_q}$. The accuracy of model $M$ can be calculated as
\begin{equation} 
    acc(M) = \frac{1}{N_q} \sum_{i=1}^{N_q} \delta(\text{IoU}(b^i_{gt}, b^i_a) > \xi) \cdot \delta(k_{gt} = k_{a}),
\end{equation}
where $N_q$ is the number of queries, $b_{gt}$ is the ground-truth bounding box, and $b_a$ is the inferred bounding box returned by model $M$. We calculate the Intersection over Union (IoU) between the answer and the ground-truth bounding boxes; the answer is correct if and only if the answer predicts the right view and the IoU is larger than $\xi = 0.5$.

\Cref{tab:cup_loc} shows the ablative study by turning on and off the joint parsing component that models human interactions, \ie, whether the model parses and tracks objects by reasoning about the interaction with agents. ``\# interactions'' means how many times the object was interacted by agents. The result shows that our system achieves an overall 88\% accuracy. Even without parsing humans, our system still possesses the ability to reason about object location by maintaining other consistencies, such as spatial consistency and appearance consistency. However, its performance drops significantly if the object was moved to different rooms. \Cref{fig:cup_false_belief} shows some qualitative results.

\setstretch{0.97}

\begin{table}[b!]
    \footnotesize
    \centering
    \caption{Accuracy of belief queries on single view subset}\label{tab:res3}
    \begin{tabular}{c|c c c}
          & True Belief & False-Belief & Overall \\ \hline
         Joint parsing acc. & 0.94 & 0.93 & 0.94 \\
         Random guessing acc. & 0.45 & 0.53 & 0.46
    \end{tabular}
\end{table}

\subsection{Experiment 2: (False-)Belief Inference}

In this experiment, we evaluate the performance of belief and false-belief inference, \ie, whether an agent's belief \ac{pg} is the same as the true object states. The evaluations were conducted on both single-view and multi-view scenarios.

\paragraph*{Multi-view} 

We collected 200 queries with ground-truth annotations that focus on the Sally-Anne false belief task. The query is defined as
\begin{equation}
    q = (k_o, t_o, b_o, k_h, t_h, b_h, t_q),
\end{equation}
where first three terms $(k_o, t_o, b_o)$ define the objects in robot $k_o$'s view located at $b_o$ at time $t_o$. Similarly, another three terms $(k_h, t_h, b_h)$ define an agent in robot $k_h$'s view located at $b_h$ at time $t_h$. The question is: where does the agent $(k_h, t_h, b_h)$ think the object $(k_o, t_o, b_o)$ is at time $t_q$?

Our system generates the answer in three steps: (i) search for the object $i$ and the agent $j$ in robot parse graphs $\tilde{pg}^{k_o}_{t_o}$ and $\tilde{pg}^{k_h}_{t_h}$, (ii) retrieve all the belief parse graphs $\bar{pg}^{k',j}_{t_q}$ at time $t_q$ to find the object $i$'s location $\bar{b}^i_{t_q}$ in human $j$'s belief, and (iii) find an object $i'$ in robot parse graph, which has the same attributes as $i$'s and has smallest distance to $\bar{b}^i_{t_q}$. The system finally returns $i'$'s location $b_{t_q}^{i'}$ as the answer.

Since there is no publicly available code on this task, we compare our inference algorithm with a random baseline model as the reference for future benchmark; it simply returns an object with the same attributes as the query object at $t_q$. The result shows that our system achieves $81\%$ accuracy, while the baseline model only has $39\%$ accuracy.

\begin{figure*}[t!]
    \centering
    \begin{subfigure}[b]{0.2\linewidth}
        \begin{overpic}[width=\linewidth]{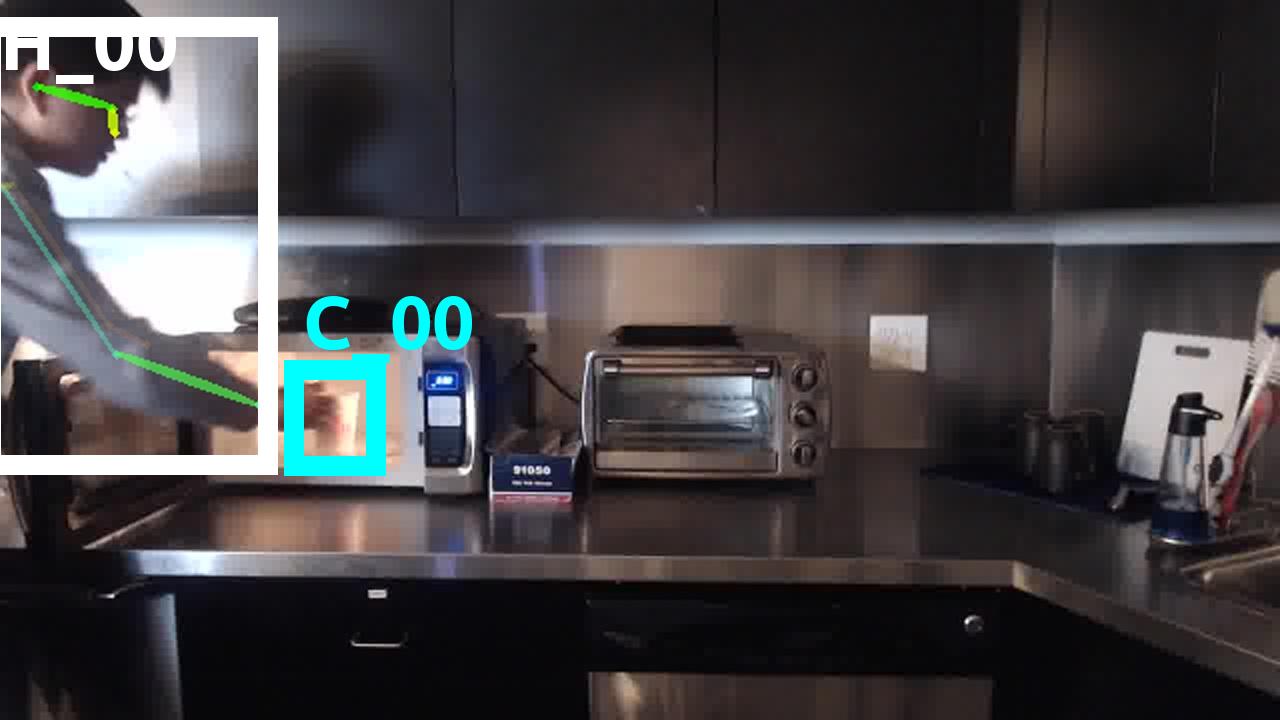}
            \put(35, 48){\color{white}w/ false-belief}
        \end{overpic}
    \end{subfigure}%
    \begin{subfigure}[b]{0.2\linewidth}
        \includegraphics[width=\linewidth]{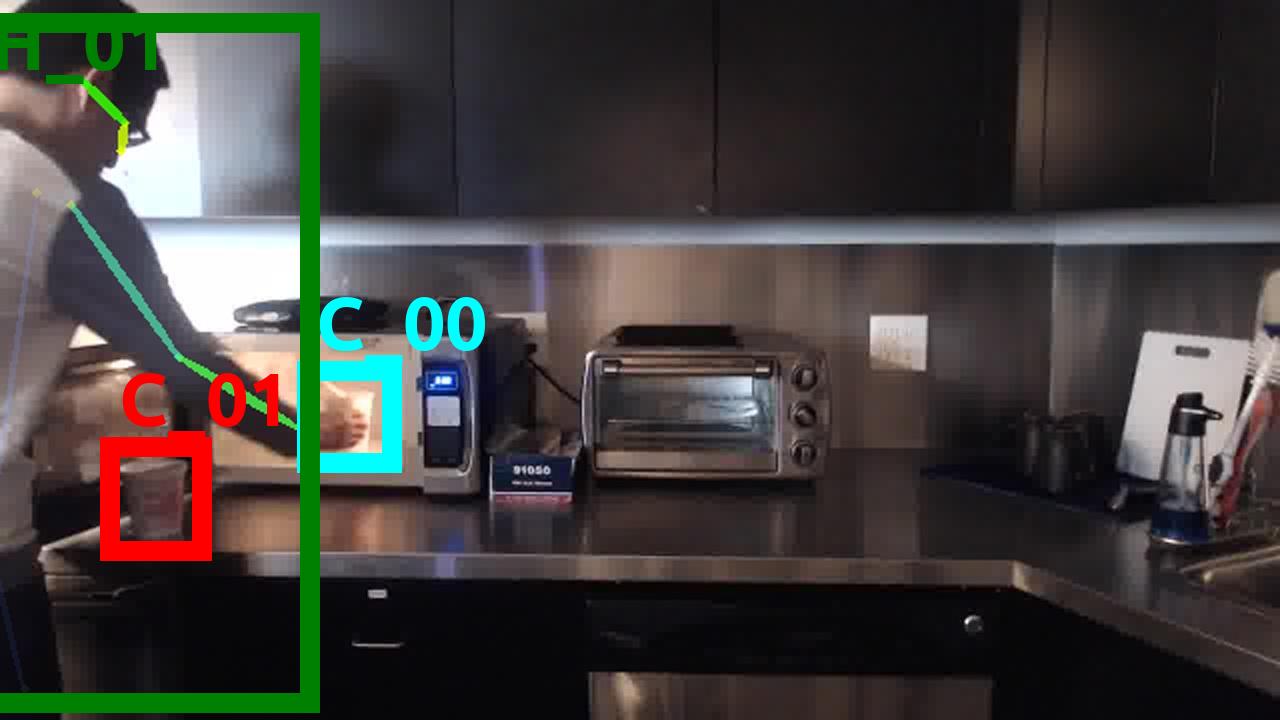}
    \end{subfigure}%
    \begin{subfigure}[b]{0.2\linewidth}
        \includegraphics[width=\linewidth]{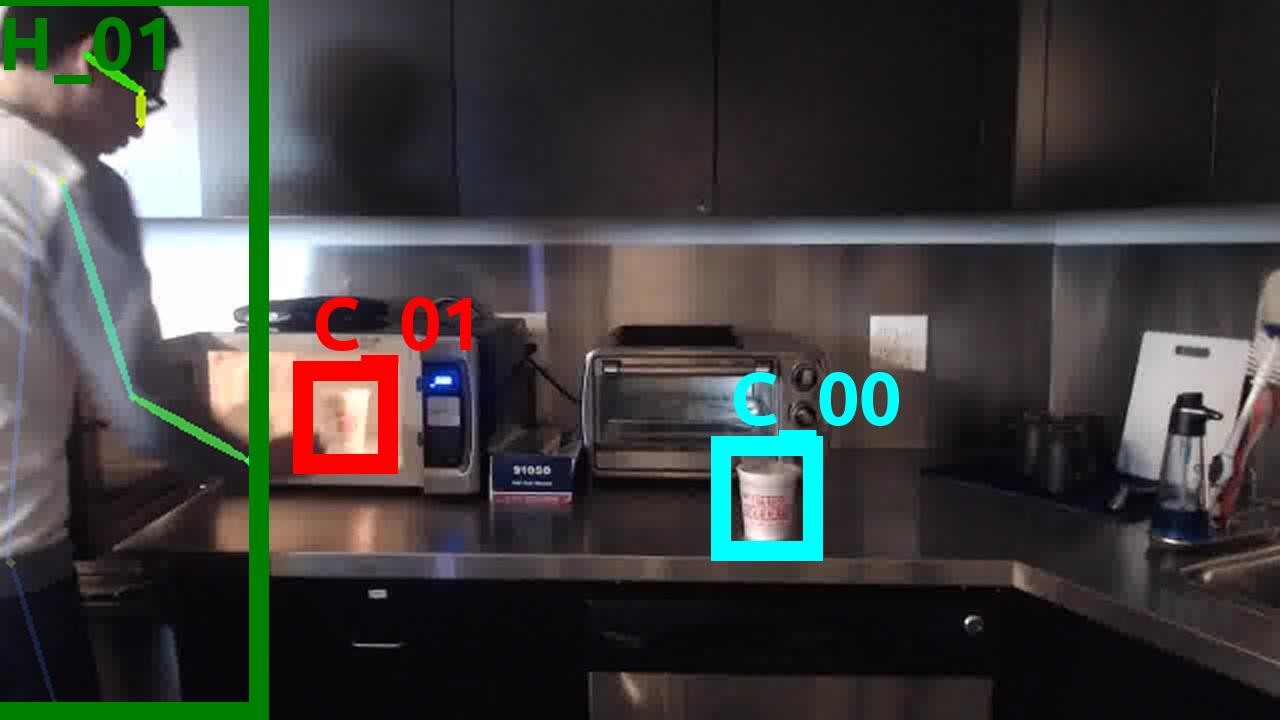}
    \end{subfigure}%
    \begin{subfigure}[b]{0.2\linewidth}
        \includegraphics[width=\linewidth]{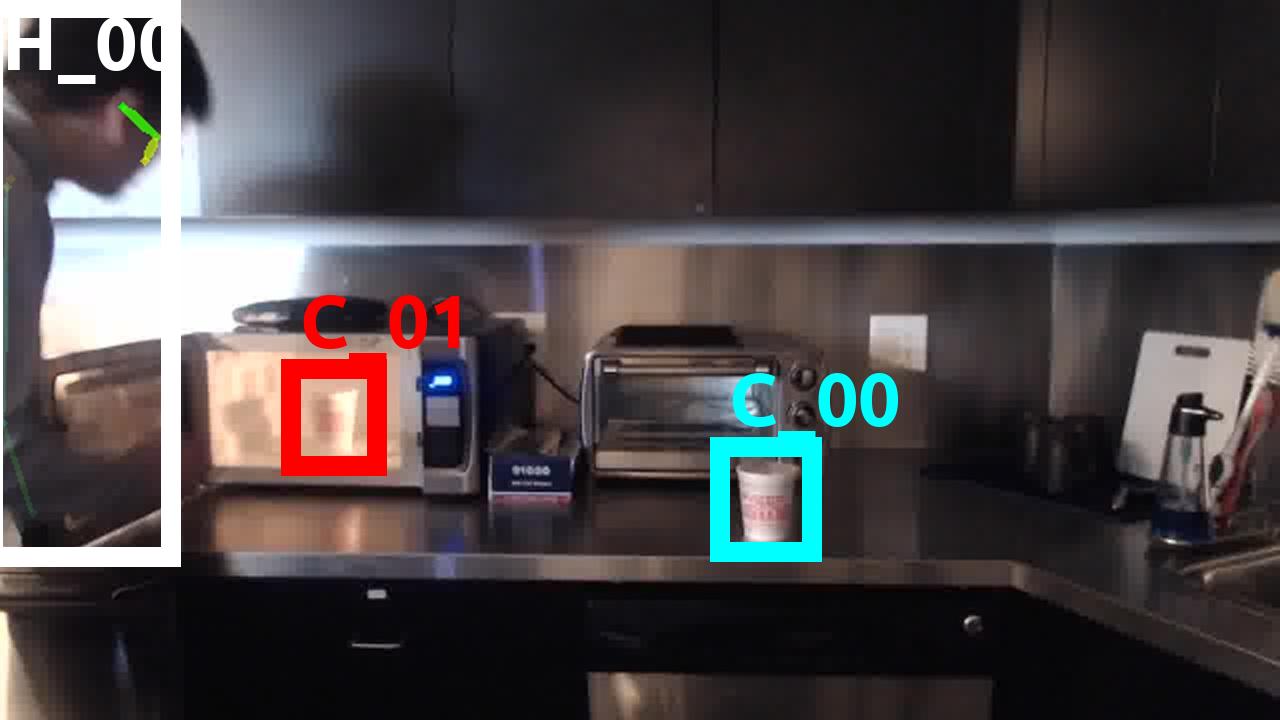}
    \end{subfigure}%
    \begin{subfigure}[b]{0.2\linewidth}
        \includegraphics[width=\linewidth]{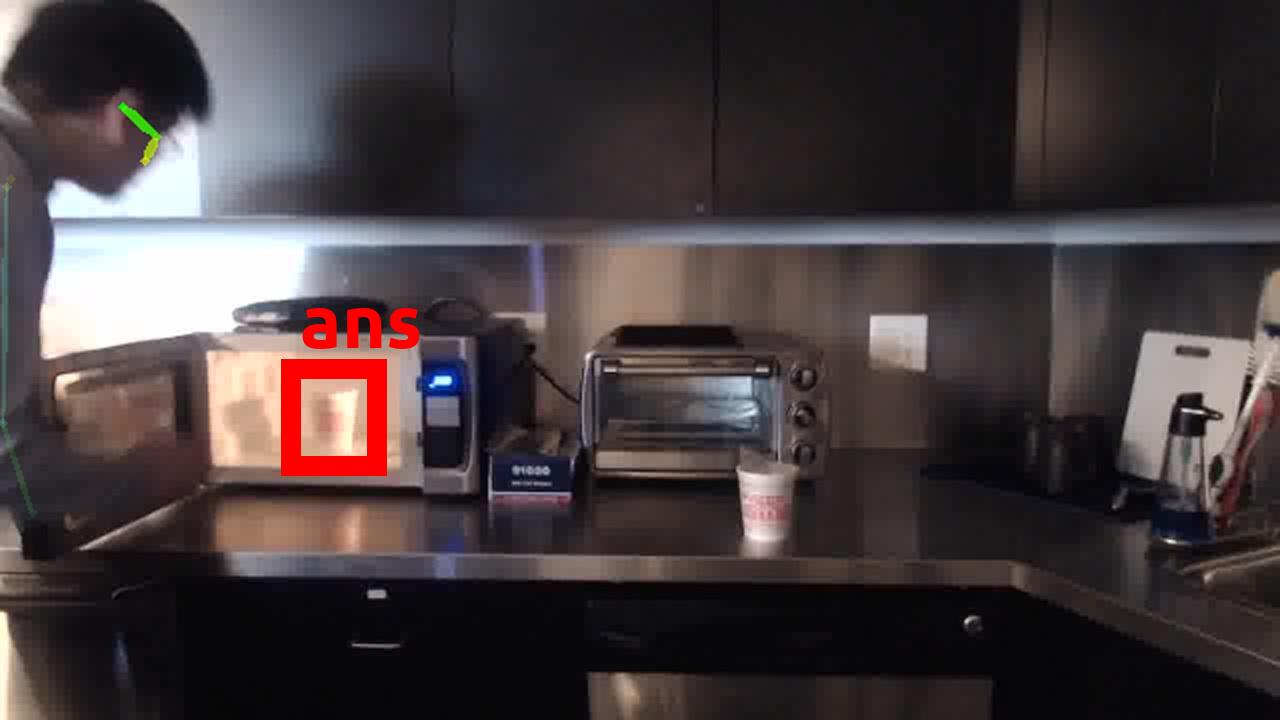}
    \end{subfigure}%
    \\
    \begin{subfigure}[b]{0.2\linewidth}
        \begin{overpic}[width=\linewidth]{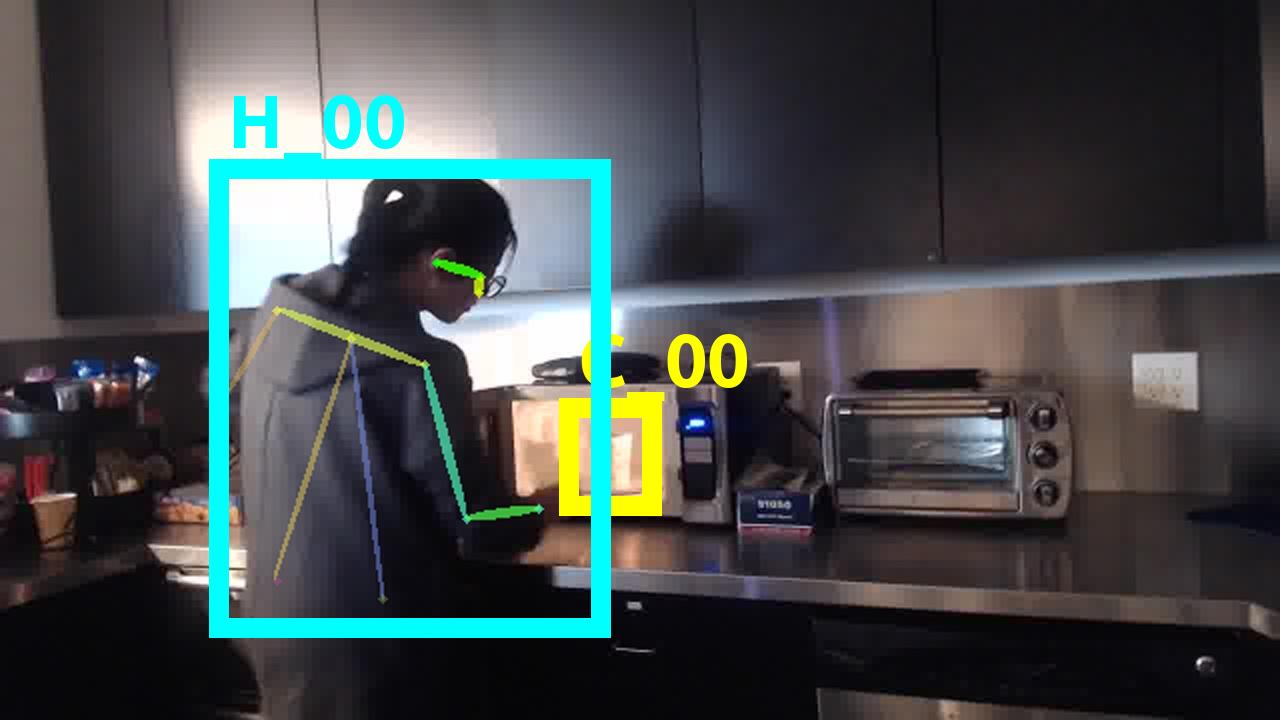}
            \put(35, 48){\color{white}w/o false-belief}
        \end{overpic}
        \caption{Observation $t_0$}
    \end{subfigure}%
    \begin{subfigure}[b]{0.2\linewidth}
        \includegraphics[width=\linewidth]{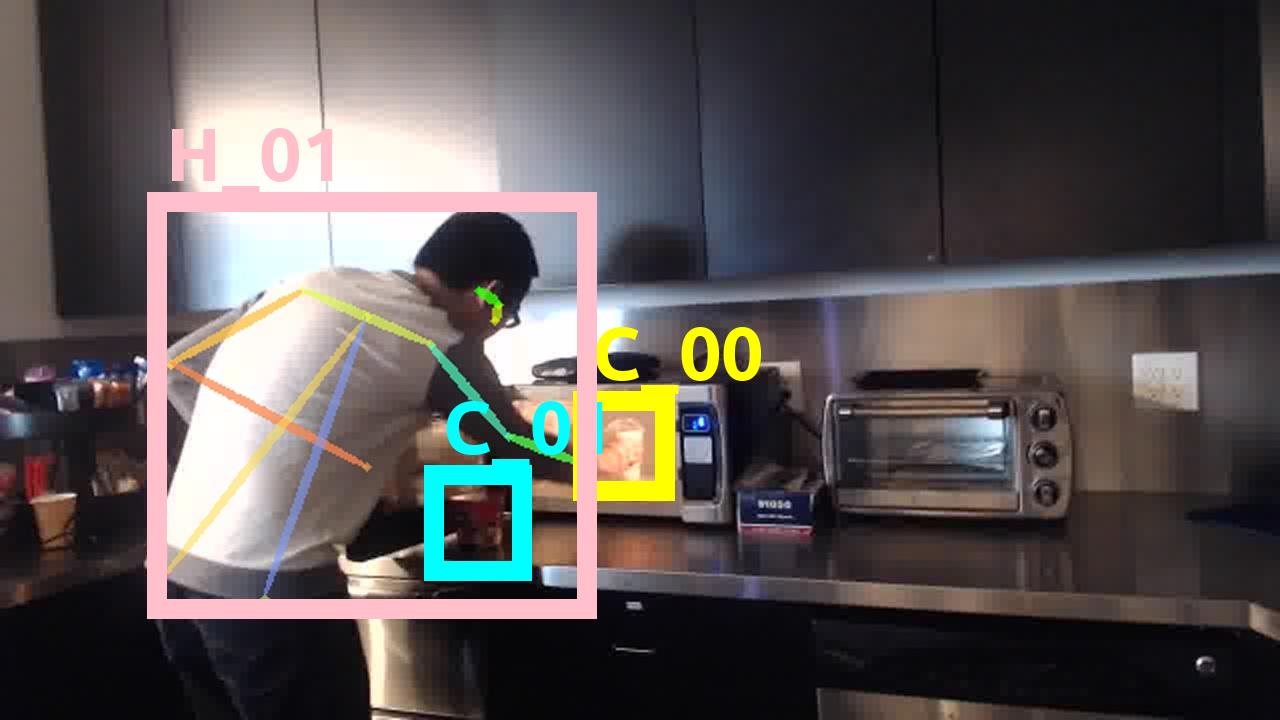}
        \caption{Observation $t_1$}
    \end{subfigure}%
    \begin{subfigure}[b]{0.2\linewidth}
        \includegraphics[width=\linewidth]{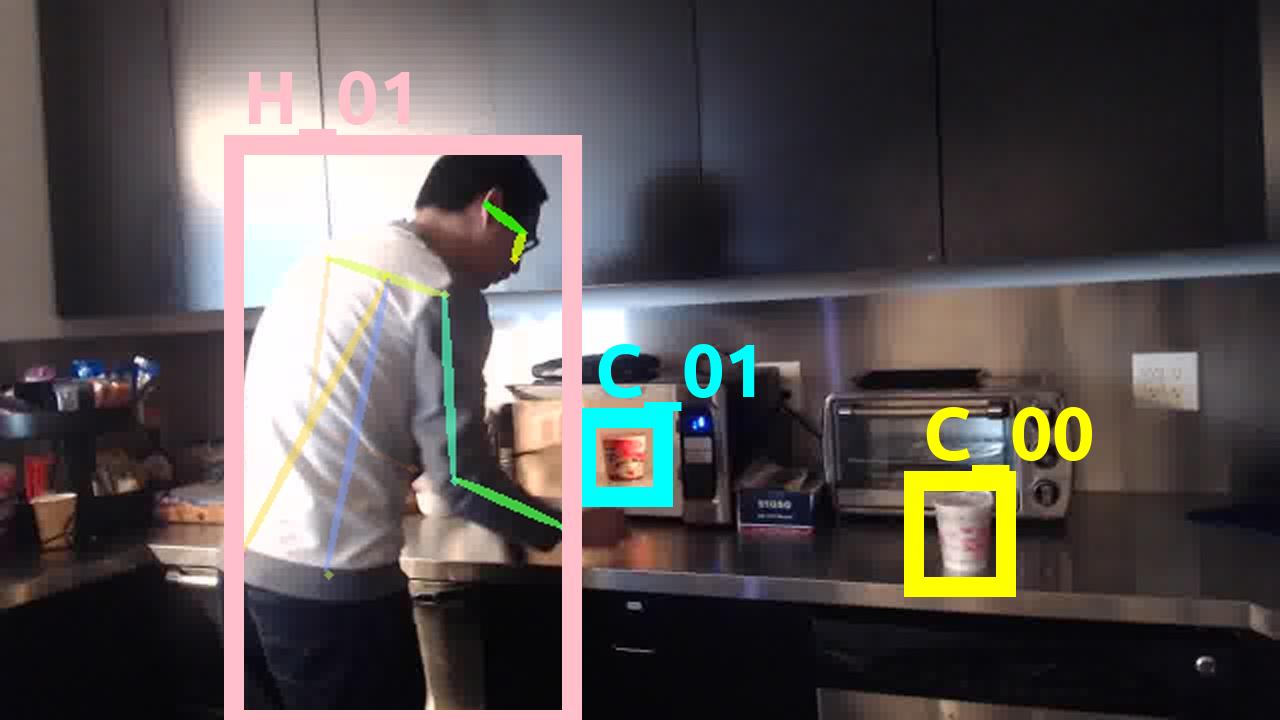}
        \caption{Observation $t_2$}
    \end{subfigure}%
    \begin{subfigure}[b]{0.2\linewidth}
        \includegraphics[width=\linewidth]{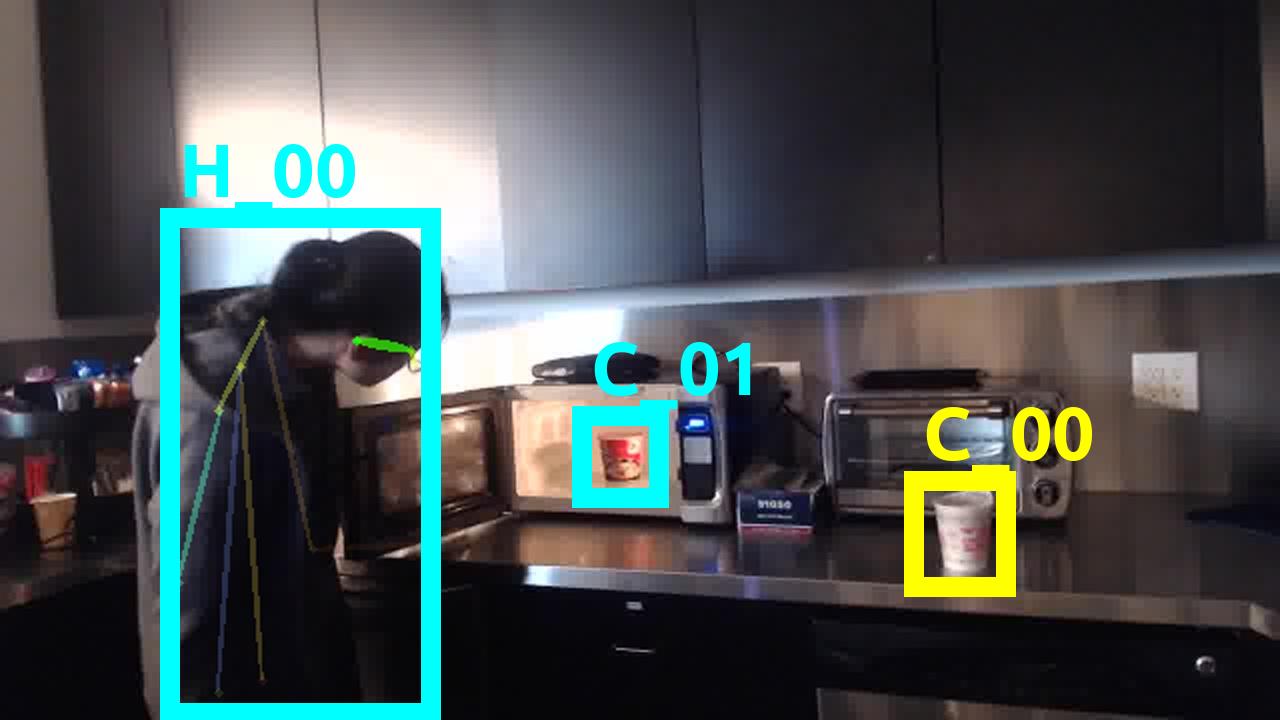}
        \caption{Observation $t_3$}
    \end{subfigure}%
    \begin{subfigure}[b]{0.2\linewidth}
        \includegraphics[width=\linewidth]{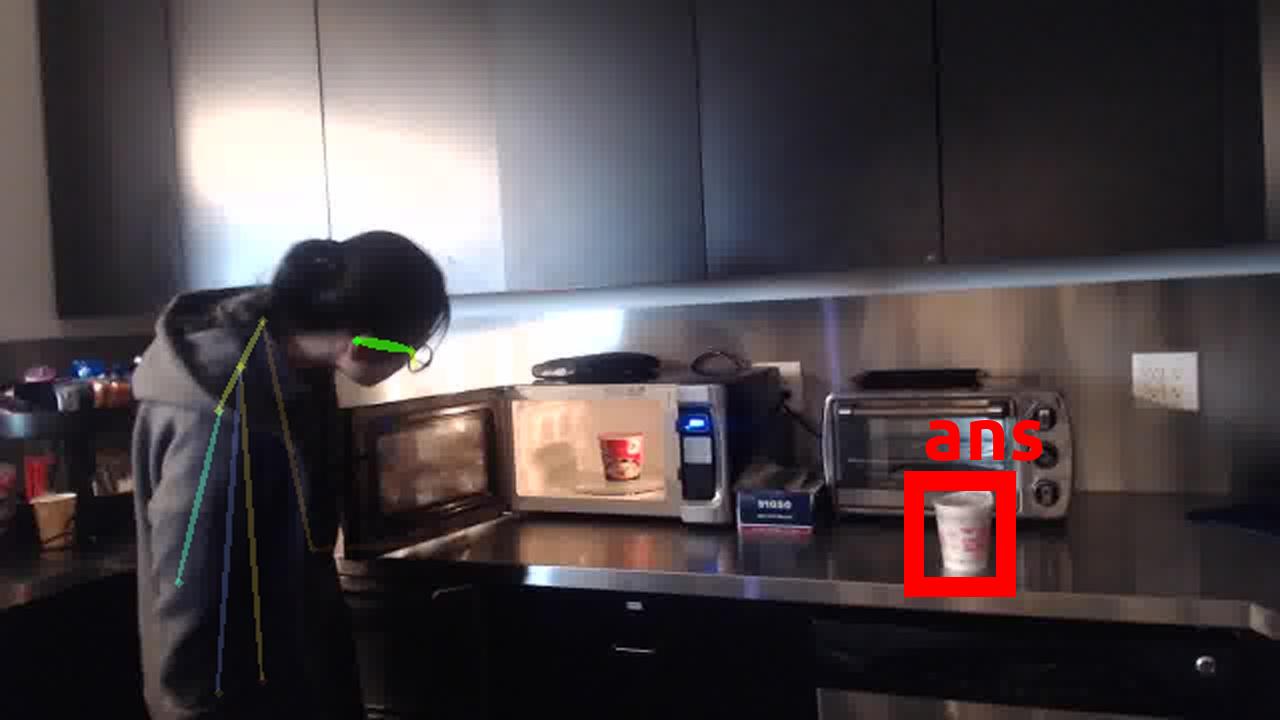}
        \caption{Prediction}
    \end{subfigure}%
    \caption{Two sample results of the Sally-Anne false-belief task in Experiment 2. (Top) with false-belief. (Bottom) Without false-belief. At time $t_0$, the first person $H_{00}$ put the cup noodle ($C_{00}$) into the microwave and leaves. The second person $H_{01}$ take out the cup noodle $C_{00}$ from the microwave at $t_1$. At time $t_2$, $H_{01}$ put his own cup noodle $C_{01}$ in the microwave and left. When $H_{00}$ returns to the room at $t_3$, our system is able to answer: where does the first person $H_{00}$ think the cup noodle $C_{00}$ is. It can successfully predict the person in the bottom row will not have the false-belief due to the different color attributes of the cups.}
    \label{fig:res1}
\end{figure*}

\begin{figure*}[t!]
    \centering
    \begin{subfigure}[b]{0.25\linewidth}
        \includegraphics[width=\linewidth]{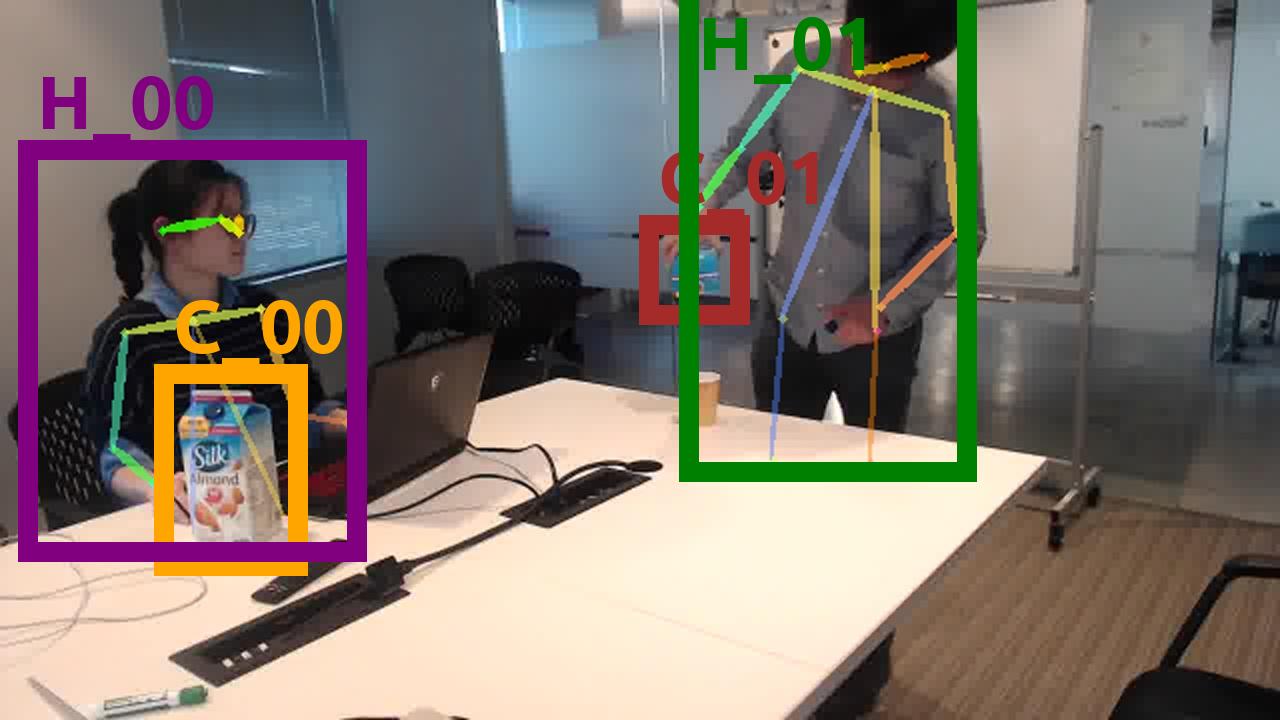}
    \end{subfigure}%
    \begin{subfigure}[b]{0.25\linewidth}
        \includegraphics[width=\linewidth]{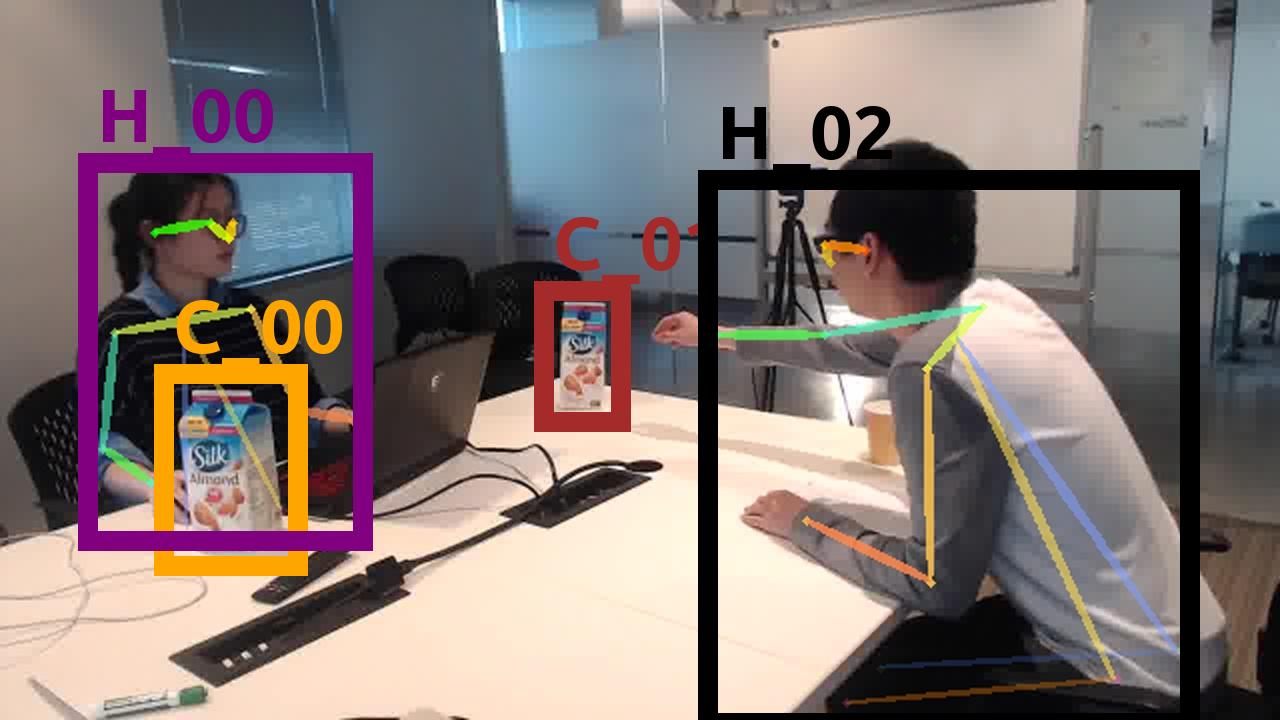}
    \end{subfigure}%
    \begin{subfigure}[b]{0.25\linewidth}
        \includegraphics[width=\linewidth]{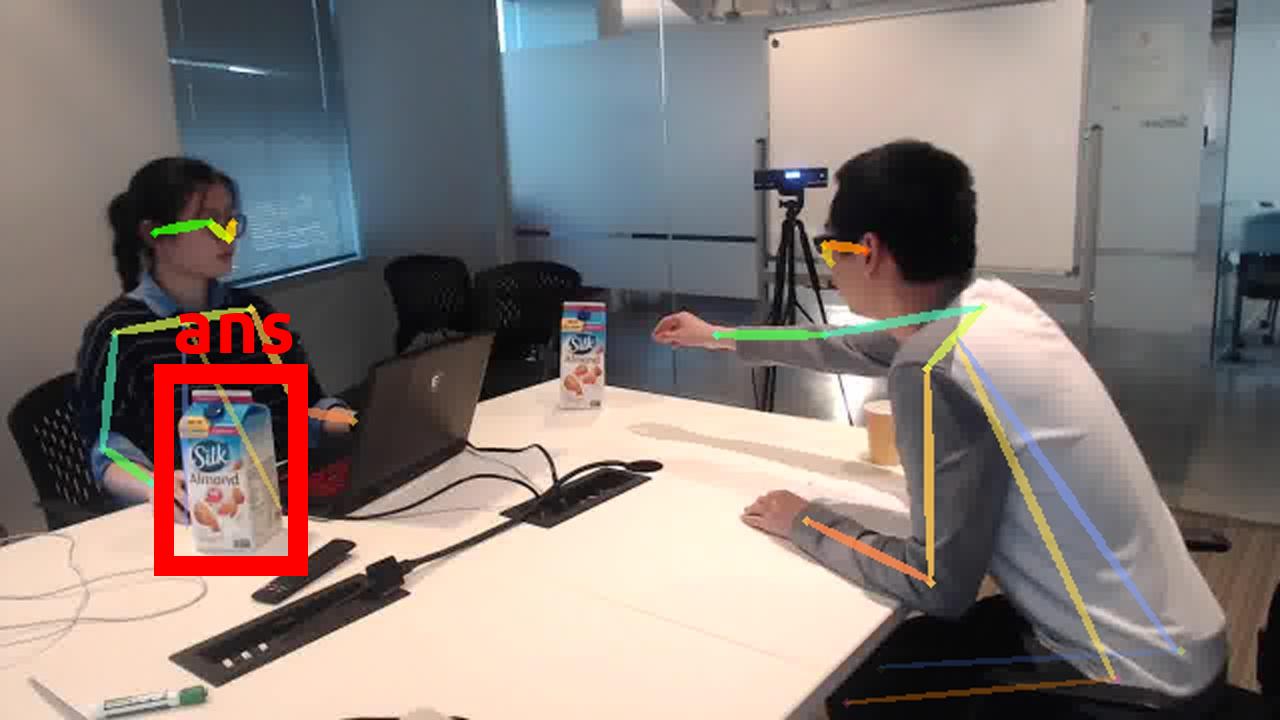}
    \end{subfigure}%
    \begin{subfigure}[b]{0.25\linewidth}
        \includegraphics[width=\linewidth]{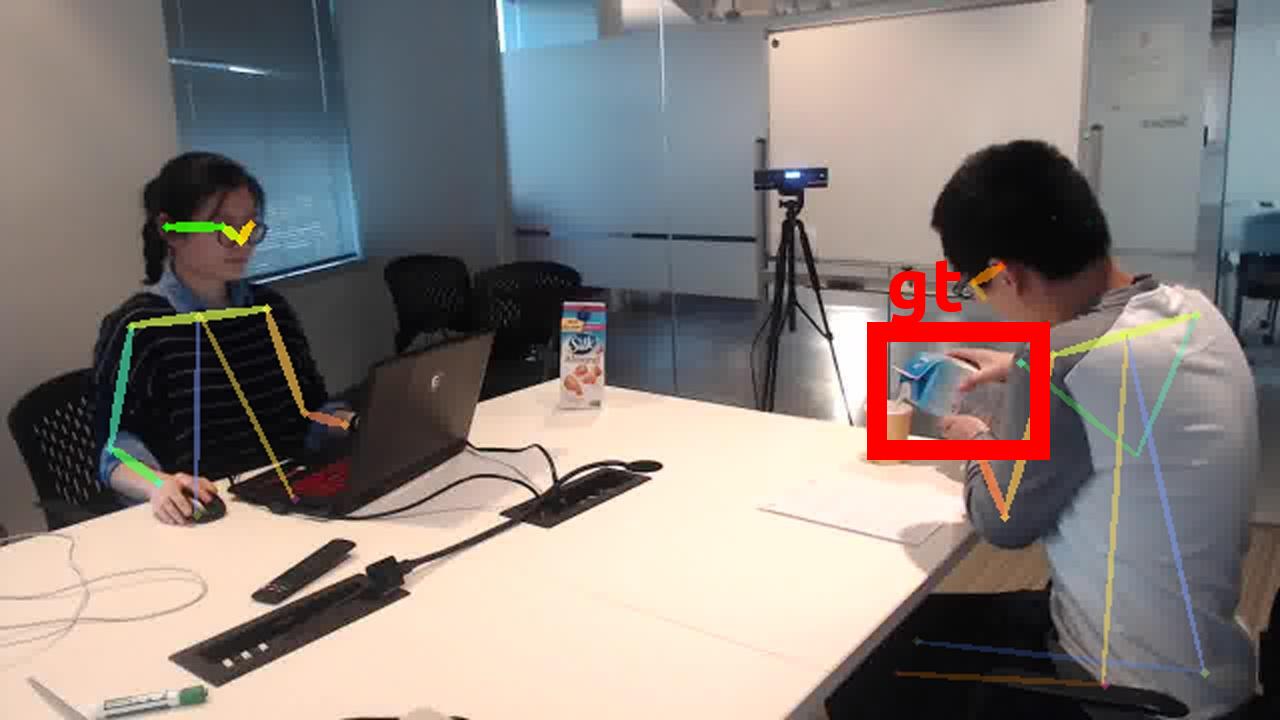}
    \end{subfigure}%
    \begin{subfigure}[b]{0.25\linewidth}
    \end{subfigure}%
    \\
    \begin{subfigure}[b]{0.25\linewidth}
        \includegraphics[width=\linewidth]{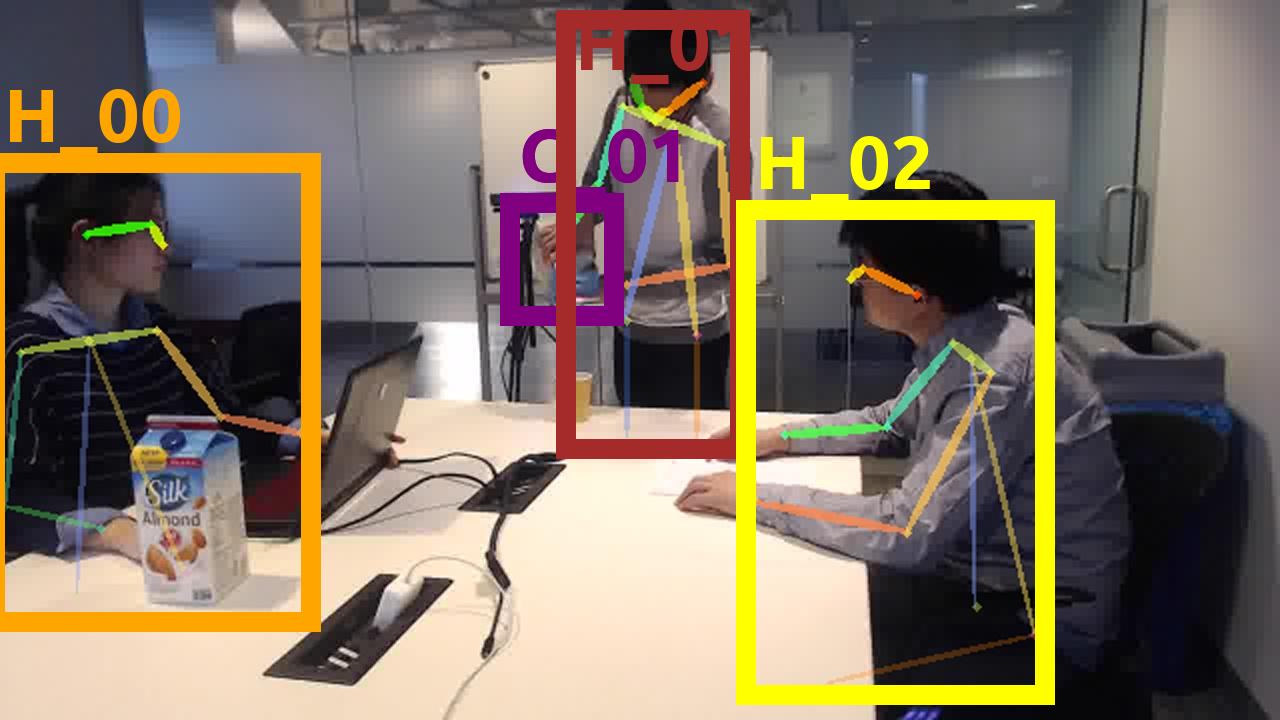}
        \caption{Observation $t_0$}
    \end{subfigure}%
    \begin{subfigure}[b]{0.25\linewidth}
        \includegraphics[width=\linewidth]{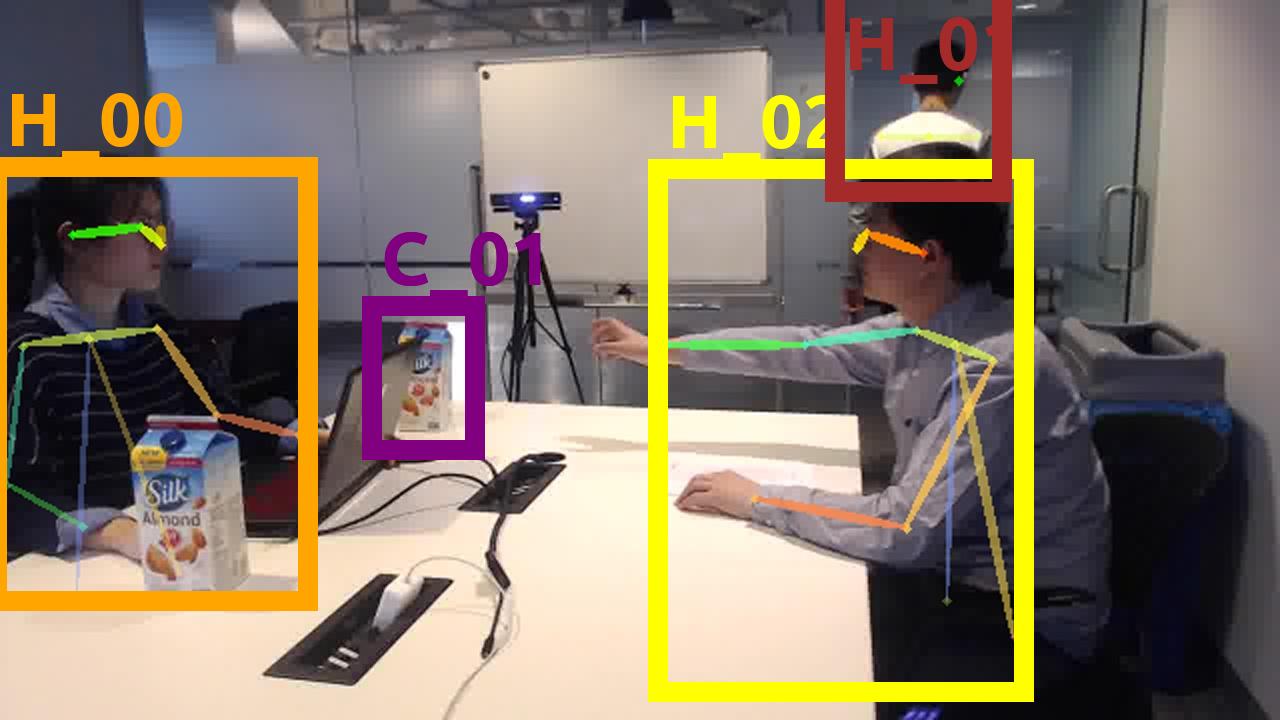}
        \caption{Observation $t_1$}
    \end{subfigure}%
    \begin{subfigure}[b]{0.25\linewidth}
        \includegraphics[width=\linewidth]{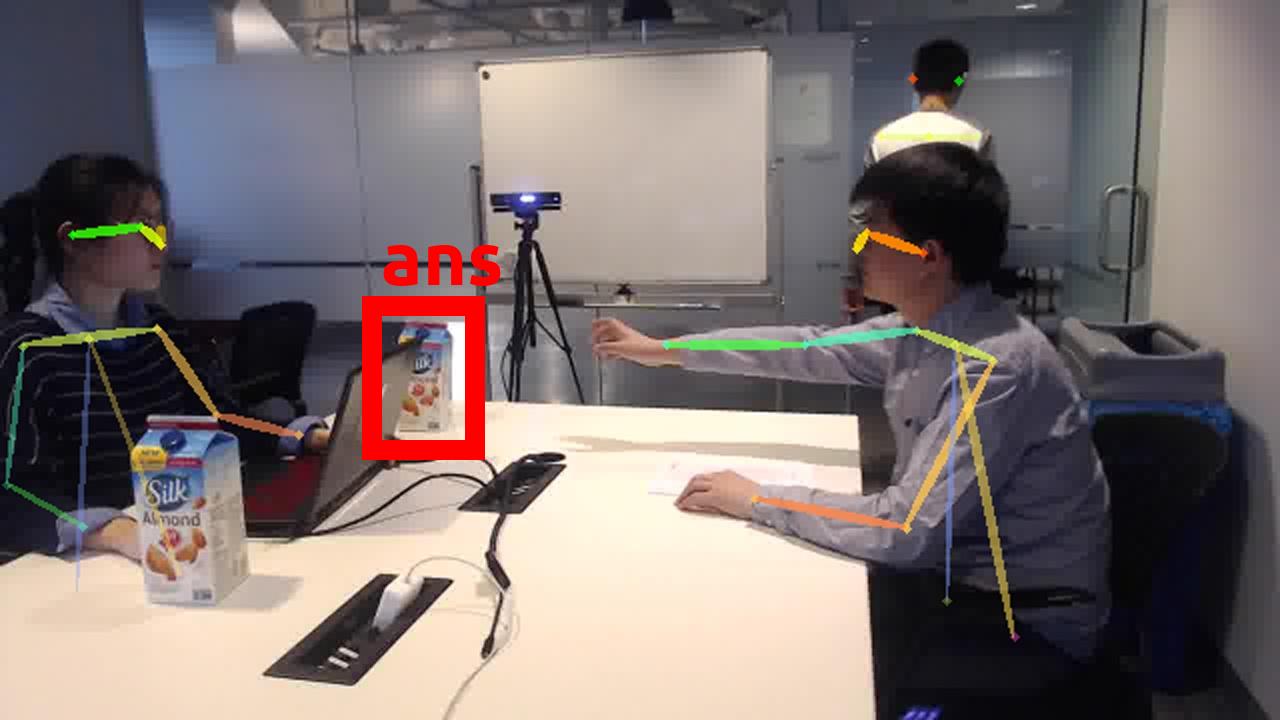}
        \caption{Prediction}
    \end{subfigure}%
    \begin{subfigure}[b]{0.25\linewidth}
        \includegraphics[width=\linewidth]{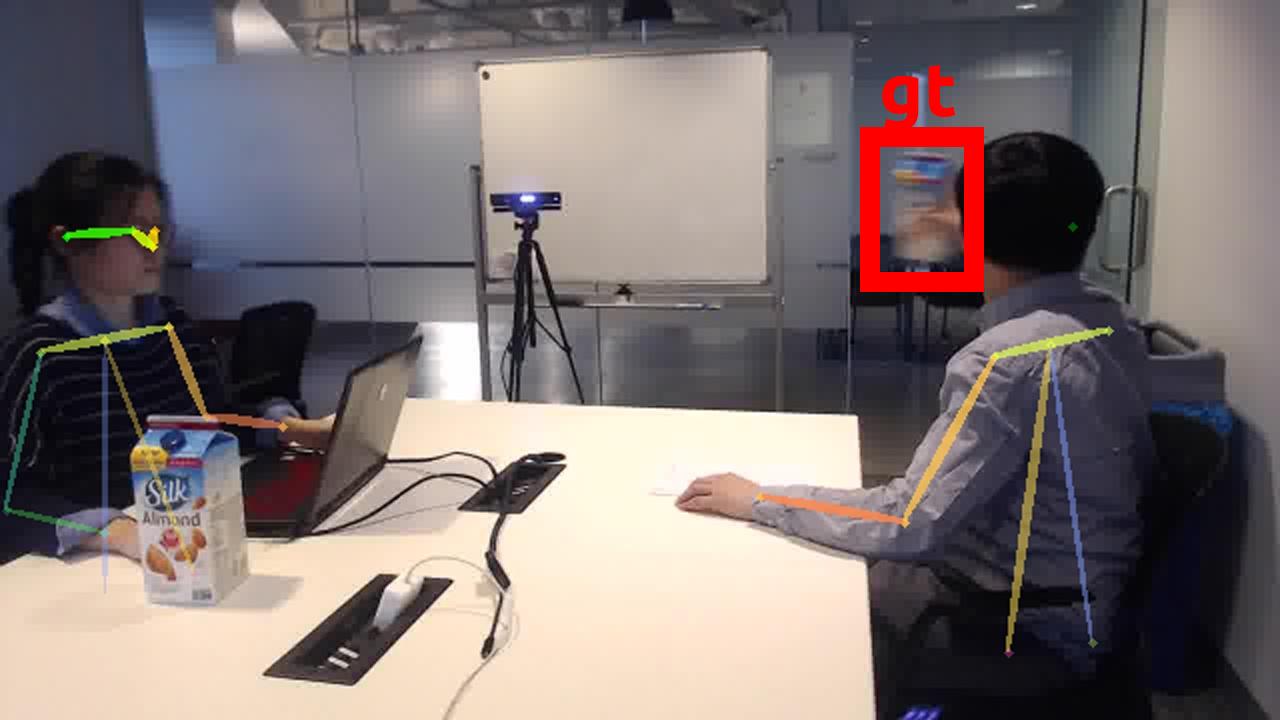}
        \caption{Ground-truth}
    \end{subfigure}%
    \caption{Two sample results of the helping task in Experiment 2. (Top) At time $t_0$, the second person ($H_{01}$) enters the room and empties the box $C_{01}$. Question: which box should the first person ($H_{00}$) give to the third person ($H_{02}$) if the first person ($H_{00}$) wants to help at $t_1$? Answer returned by the system correctly infers that $H_{02}$ has a false-belief, and $H_{00}$ should give another box ($C_{00}$) to $H_{02}$, rather than the $C_{01}$ his is reaching for. (Bottom) Since the person $H_{02}$ observes the entire process, there is no false-belief; in this case, $H_{02}$ reaches for $C_{01}$ is to throw it away.}
    \label{fig:res2}
\end{figure*}

\paragraph*{Single-view}

We collected a total of 100 queries, including two types of belief inference tasks: the Sally-Anne false-belief task and the helping task, as shown in \cref{fig:sally}. The queries have two forms
\begin{equation}
    q = (t_o, b_o, b_h, t_q), \quad \emph{and} \quad
    q = (b_h, t_q),
\end{equation}
indicating two different types of questions: (i) where does the agent $b_h$ think the object $(t_o, b_o)$ is at time $t_q$? (ii) Which object will you give to the agent $(t_q, b_h)$ at time $t_q$ if you would like to help? For the first type of questions, \ie, the Sally-Anne false-belief task, similar to the multi-view setting, the system should return the object bounding box as the answer. For the second type of question, \ie, the helping task, the system first infers whether the agent has false-belief. If not, the system returns the object the person wants to interact based on their current pose; otherwise, the system returns another suitable object closest to them. Qualitative results are shown in \cref{fig:res1,fig:res2}, and quantitative results are provided in \cref{tab:res3}.

\section{Conclusion and Discussions}\label{sec:conclusion}

In this paper, we describe the idea of using \ac{pg} as a unified representation for tracking object states, accumulating robot knowledge, and reasoning about human (false-)beliefs. Based on the spatiotemporal information observed from multiple camera views of one or more robots, robot \ac{pg} and belief \ac{pg} are induced and merged to a joint \ac{pg} to overcome the possible errors originated from a single view. With such a representation, a joint inference algorithm is proposed, which possesses the capabilities of tracking small occluded objects across different views and inferring human (false-)beliefs. In experiments, we first demonstrate that the joint inference over the merged \ac{pg} produced better tracking accuracy. We further evaluate the inference on human true- and false-belief regarding objects' locations by jointly parsing the \ac{pg}s. The high accuracy demonstrates that our system is capable of modeling and understanding human (false-)beliefs, with the potential of helping capability as demonstrated in developmental psychology.

\ac{tom} and Sally-Anne test are interesting and difficult problems in the area of social robotics. For a service robot to interact with humans in an intuitive manner, it must be able to maintain a model of the belief states of the agents it interacts with. We hope the proposed method using a graphical model has demonstrated a different perspective compared to prior methods in terms of flexibility and generalization. In the future, a more interactive and active set up would be more practical and compelling. For instance, by integrating activity recognition modules, our system should be able to perceive, recognize, and extract richer semantic information from the observed visual input, thereby providing more subtle (false-)belief applications. Communication, gazes, and gestures are also crucial in intention expression and perception in collaborative interactions. By incorporating these essential ingredients and taking advantage of the flexibility and generalization of the model, our system should be able to go from the current passive query to active response to assist agents in real-time.

\setstretch{1}

\end{document}